\documentclass[journal]{IEEEtran}

\ifCLASSINFOpdf
\else
\fi
\usepackage{times} 
\usepackage{xcolor}
\usepackage{soul}
\usepackage[utf8]{inputenc}
\usepackage[small]{caption}
\usepackage{multirow}
\usepackage{graphicx}  
\usepackage{helvet}
\usepackage{courier}
\usepackage{url}
\usepackage{float}
\usepackage{color}
\usepackage{comment}
\usepackage{amsmath}
\usepackage{amssymb}
\usepackage{booktabs}
\usepackage{verbatim}
\usepackage{amsthm}
\usepackage{url}
\usepackage[ruled,vlined]{algorithm2e}

\usepackage[loose]{subfigure}
\usepackage{epstopdf}

\usepackage{float}

\def\etal{{\em et al.}}

\begin{document}
	%
	\title{Latent Heterogeneous Graph Network for Incomplete Multi-View Learning}
	%
	
	\author{Pengfei Zhu, Xinjie Yao, Yu Wang, Meng Cao, Binyuan Hui, Shuai Zhao, Qinghua Hu
		\thanks{Pengfei Zhu, Xinjie Yao, Yu Wang, Meng Cao, Binyuan Hui, Qinghua Hu are with College of Intelligence and Computing, Tianjin	University, Tianjin, 300072 China.}
		\thanks{Shuai Zhao is with the College of Intelligence and Computing, Tianjin	University, Tianjin, China, and is also with the Automotive Data of China (Tianjin) Co., Ltd, Tianjin, China.}
		\thanks{Yu Wang (wangyu\_@tju.edu.cn) is the corresponding author.}
		\thanks{
		
	}}
	
	\markboth{IEEE TRANSACTIONS ON MULTIMEDIA}%
	{Shell \MakeLowercase{\textit{et al.}}: Bare Demo of IEEEtran.cls for IEEE
		Journals}
	
	\maketitle
	\begin{abstract}
		Multi-view learning has progressed rapidly in recent years. Although many previous
		studies assume that each instance appears in all views, it is common in real-world applications for instances to be missing from some views, resulting in incomplete multi-view data. To tackle this problem, we propose a novel Latent Heterogeneous Graph Network (LHGN) for incomplete multi-view learning, which aims to use multiple incomplete views as fully as possible in a flexible manner. By learning a unified latent representation, a trade-off between consistency and complementarity among different views is implicitly realized. To explore the complex relationship between  samples and latent representations, a neighborhood constraint and a view-existence constraint are proposed, for the first time, to construct a heterogeneous graph. Finally, to avoid any inconsistencies between training and test phase, a transductive learning technique is applied based on graph learning for classification tasks. Extensive experimental results
		on real-world datasets demonstrate the effectiveness of our model over existing state-of-the-art approaches.
	\end{abstract}
	
	\begin{IEEEkeywords}
		Incomplete multi-view learning, heterogeneous
		graph network, graph learning.
	\end{IEEEkeywords}

	\section{Introduction}
	In real-world applications, data are usually represented by different views given by various modalities or types of features \cite{2019Jointly,2015Multi}. 
	For example, a web document can be represented by its URL and by the words on the page; images can be described by a variety of visual features such as GIST \cite{OlivaT01}, Gabor \cite{LadesVBLMWK93},
	HOG \cite{hog}, and SIFT \cite{sift}; and film segments can be represented by video
	features and voices. Such forms of data are referred to as multi-view
	data, which have been attracted considerable interest by many researchers. Multi-view learning (MVL) aims to exploit the consistency and complementarity of information obtained from different views \cite{ZhaoDF17}. By using the complementary characteristics among different views, MVL can achieve better performance than that from just a single view \cite{Sun13}.
	
\begin{figure}[t]
		\centering
		\includegraphics[width=1\linewidth]{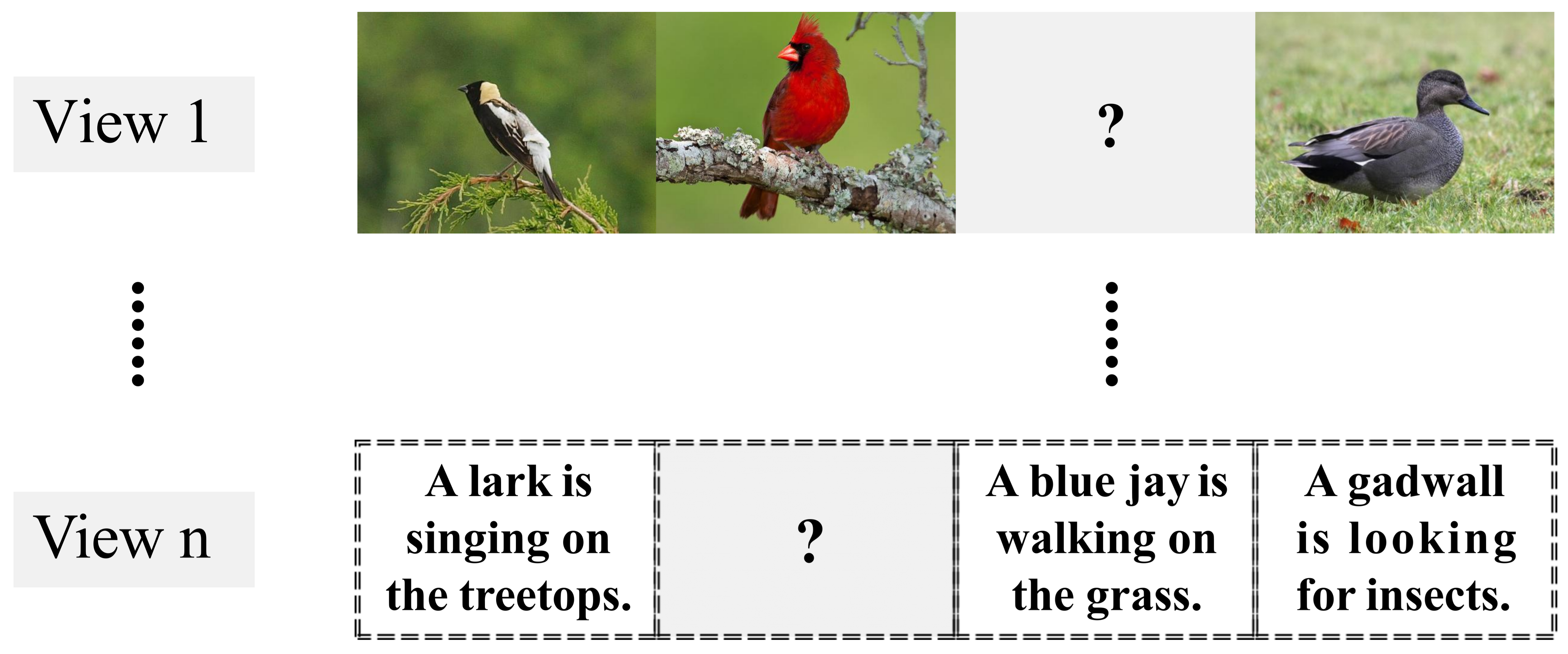}
		\caption{An example of incomplete multi-view data. A sample can be represented by different modalities such as images, text, and so on. Several uncontrollable factors mean that some views may be missing.}
		\label{IMVD1}
	\end{figure}

	Most traditional multi-view researches assume that all the examples have complete information from all views, i.e., each example of the data has a complete feature set \cite{Ding-et-al:2016,xu:mvl,NieCLL18,8587193}.
	However, in real-world applications, several uncontrollable factors mean that the collected multi-view data are usually incomplete \cite{LiZHZW19,LiuWZLZLLDY20}. For example, in Fig.~\ref{IMVD1}, the birds can be represented by visual and textual features. Some samples only have visual or text information, and only some of them share these features. We call such a dataset ``incomplete multi-view data.'' These missing views lead to many problems in MVL. First, the balanced information of multiple views is seriously broken because these views may have different numbers of instances and features. Second,  it is difficult to find complementary and consistent information for these incomplete views. The violation of this assumption makes incomplete MVL (\textbf{IMVL}) a very challenging task.
	
	Recently, many IMVL methods have been proposed, and can be divided into three types: data imputation-based methods, data grouping-based methods, and elastic representation-based methods. Data imputation-based methods first fill the missing views, and then apply a common MVL algorithm \cite{liu2018late,2020Adaptive,Lin_2021_CVPR,Tran0ZJ17}. Some widely used filling methods include zero filling, mean value filling, and k-nearest neighbor filling \cite{MarlinZRS11}. Although this type of method is effective in some cases, however, it relies on a large number of paired data, making this approach very expensive and restricted in practical applications. At the same time, harmful estimation noise may be included when estimating the missing view, and the introduction of such virtual samples may have a negative impact on the model. Another natural strategy is to first manually group samples according to the availability of data,  and then to train multiple models on these
	groups for subsequent fusion. Although grouping-based methods are more effective than those attempt to learn each individual view, it may result in less available data, which can easily lead to over-fitting. Additionally, grouping-based methods are relatively inflexible, especially for data with a large number of views. Elastic representation-based methods can effectively avoid the limitations of the other methods, but the current elastic representation mainly focuses on the use of completeness theory to learn better latent representations \cite{WangZLYZ19}, while ignoring the complex relationship between samples. 
	
	Although existing methods provide some schemes to address the IMVL problem, they still suffer from the following issues: (1) Most of the previous methods use the shallow models to obtain the common representation, but do not fully exploit the complex interaction between views and samples from incomplete multi-view data. (2) Few of them consider the negative effect of the inconsistency between training and test phases. 
Since the missing data are randomly distributed between the training
	and test sets, there may be inconsistencies between training and test. 
	
	To solve these challenges, we propose an effective and flexible IMVL network called the Latent
	Heterogeneous Graph Network (LHGN). First, information of different views is encoded into a unified latent representation that can implicitly achieve the best trade-off between consistency and complementarity for different views.
	To mine the structural information of the samples, we propose, for the first time, a view-existence constraint and a neighborhood constraint, which are used to construct a heterogeneous graph on the unified latent representation. Then, graph learning, concretely, the graph attention convolution network (GAT), is applied to take advantage of the complementary structural information between samples based on the constructed graph. Finally, the aggregation network is used to fuse the features of meta-paths from different views, allowing different downstream tasks to be performed on the aggregated features.
	\begin{figure*}[t]
		\centering
		\includegraphics[width=1\linewidth]{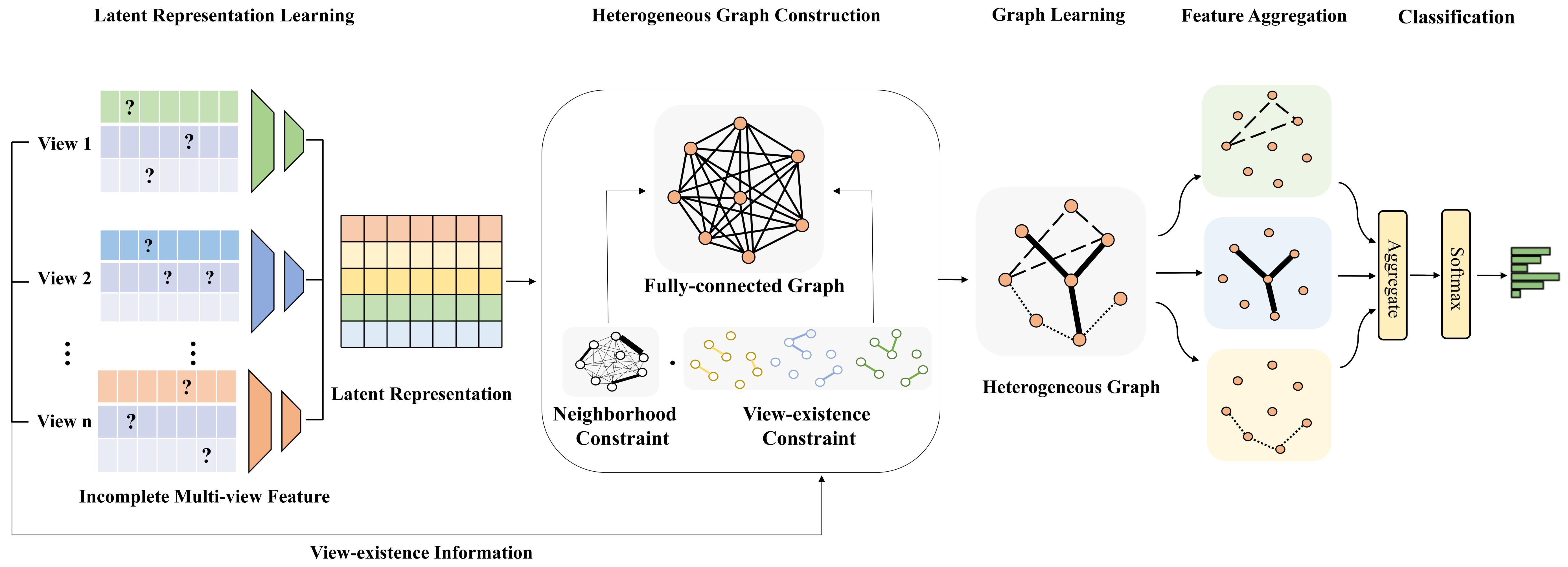}
		\caption{The network architecture of Latent Heterogeneous Graph Network (LHGN) for incomplete multi-view
			learning. LHGN consists of five parts, i.e., several view-specific encoder networks that learn a common latent representation of different views, a latent heterogeneous graph construction layer, a graph learning layer, an aggregation layer that concatenates the learned features, and a classification layer.
}
		\label{framework}
	\end{figure*}
	The main contributions of this paper can be summarized as follows:
	
	\begin{itemize}
		\item We propose a neighborhood constraint and a view-existence constraint for the IMVL problem. Based on the constraints, we construct a heterogeneous graph from the latent representation that can flexibly handle all kinds of incomplete cases. Through heterogeneous graph learning, the complex relationship among samples in the hidden space can be fully exploited.
		\item To avoid inconsistencies between the training and the test process, we exploit the method of transductive learning to construct a heterogeneous graph of the training samples and test samples. Using graph learning, the representation of the test samples can be updated during the training process. When the training process is complete, the learned representation of the test samples can be directly used for downstream tasks.
		
		\item Experiments on nine real-world multi-view datasets demonstrate the superiority of LHGN over other state-of-the-art approaches, and show that it is robust to view missing.
	\end{itemize}
	
	The remainder of this paper is organized as follows. Section II introduces some
	related work on MVL, IMVL, and
	graph neural networks. In Section III, we describe the proposed method in detail.
	Experimental results on nine incomplete multi-view datasets are presented in
	Section IV. Finally, the conclusions of this study are summarized in Section V.
	
	\section{Related Work}
	
	\label{s2}
	\subsection{Multi-view Learning}
	The core of MVL is to use the consistency and complementarity among different views to discover the underlying pattern of the data \cite{yan2021deep}. Existing MVL methods can be divided into three groups: (1) co-training, (2) canonical correlation analysis (CCA), and (3) multiple kernel learning (MKL).  Co-training algorithms operate under the consideration of a multiple views consensus. The main aim of co-training is to maximize the mutual agreement across all views and reach the broadest consensus \cite{KumarD11,AppiceM16,SunLXB15,KumarRD11}.
	CCA-based multi-view models have been widely used for MVL. CCA \cite{cca} seeks a common representation by maximizing the correlation between different views. Representative algorithms include deep CCA (DCCA) \cite{dcca} and the deep canonical correlation autoencoder (DCCAE) \cite{dccae}. MKL algorithms are developed to boost the search space capacity of possible kernel functions, e.g., linear, polynomial, and Gaussian kernels, in an attempt to achieve good generalization. Representative MKL algorithms include semidefinite programming \cite{LanckrietCBGJ03}, SMO \cite{BachLJ04}, semi-infinite linear programming \cite{SonnenburgRSS06}, SimpleMKL \cite{SimpleMKL} and JLMVC \cite{2019Jointly}.

	\subsection{Incomplete Multi-view Learning}
	Most existing MVL methods assume that all instances
	are present in all views. However, this assumption does not always hold in real-world applications. Recently, many IMVL methods have been proposed \cite{2018Late}, and can be roughly categorized into three types: (1) data imputation-based methods, (2) data grouping-based methods, and (3) elastic representation-based methods. Data imputation-based methods focus on imputing the missing data and then use the common MVL methods on the imputed completed data. Xu et al. \cite{comlet1} and Shao et al. \cite{comlet2} collectively completed the kernel matrices of incomplete views by optimizing the alignment of shared instances. However, their methods cannot be extended to widely used multi-view methods. To address this issue, Tran et al. \cite{Tran0ZJ17} proposed a novel Cascaded Residual Autoencoder (CRA) to impute missing modalities. By stacking residual autoencoders, CRA grows iteratively to model the residual between the current prediction and original data. 
	
	Data grouping-based methods aim to group the data according to the existence of data and then divided it into multiple learning tasks. For example, Yuan et al. \cite{YuanWTNY12} proposed to divide the samples according to the availability of data sources, and then to learn shared sets of features with state-of-the-art sparse learning methods. Li et al. \cite{pvc} proposed to use information about the instance alignment to learn a common latent subspace for aligned instances and a private latent representation for unaligned instances via NMF. Following this approach, various extensions have been proposed \cite{img,gans}. However, these methods fail to capture the hidden distribution of missing data flexibly. 
	
	To this end, some elastic representation-based methods have been proposed to deal with this issue. For instance, Shao et al. \cite{group1} introduced a novel graph Laplacian term to couple the complete samples in the latent subspace, thus making similar samples more likely to be grouped together. Yin et al. \cite{latent1} proposed a regression-like objective for learning a latent representation, while Zhao et al. \cite{latent2} developed a deep semantic mapping method for coupling incomplete multi-view samples. Zhang et al. \cite{cpm} proposed an elastic latent space representation method that learns a complete common representation, enabling the complex relationship between multiple samples to be mined.
	
	\subsection{Heterogeneous Graph Neural Networks}

	\textbf{Heterogeneous Graph.} A heterogeneous graph
	\cite{Heterogeneous}, denoted as $\mathcal{G}=(\mathcal{V},
	\mathcal{E})$, consists of a sample set $\mathcal{V}$ and an edge set
	$\mathcal{E}$. A heterogeneous graph is associated with a node-type mapping function $\phi: \mathcal{V}
	\rightarrow
	\mathcal{A}$ and a link-type mapping function $\psi: \mathcal{E}
	\rightarrow
	\mathcal{R}$. $\mathcal{A}$ and $\mathcal{R}$ denote the sets of
	predefined
	object types and link types, where $|\mathcal{A}|+|\mathcal{R}|>2$.

	Heterogeneous graphs with different types of nodes and edges are
	ubiquitous in real life, and have significant values in many applications.
	Heterogeneous graph networks use the rich semantic association between
	different types of edges hidden in the graph to pursue a meaningful vector
	representation for each node \cite{ChangHTQAH15, ZhangXKLMZ18}. Graph
	neural networks (GNNs) employ deep neural networks to aggregate feature
	information from neighboring nodes, which makes the aggregated embedding more
	powerful. Recently, a handful of studies have focused on heterogeneous graph
	embedding using GNNs. The basic idea of these models is to
	split a heterogeneous graph into multiple homogeneous subgraphs. For
	instance, GCN \cite{KipfW17}, GAT \cite{gat}, and GraphSAGE \cite{graphsage}
	employ a convolutional operator, a self-attention mechanism, and a long short-term memory
	architecture, respectively, to aggregate the feature information of neighboring nodes. HAT \cite{hat} uses different meta-path-defined edges to
	extract subgraphs, and then embeds heterogeneous graphs with the attention
	mechanism of GNNs. RGCN \cite{SchlichtkrullKB18} retains a
	distinct linear projection weight for each edge type to deal with the
	highly multi-relational data. HetGNN \cite{zhang2019heterogeneous} adopts
	different recurrent neural networks for encoding the heterogeneous content of nodes and type-based
	neighbor aggregation, and employs heterogeneous types to integrate multimodal
	features.

	\section{Proposed Method}
	As described aforementioned, although existing IMVL methods can obtain high-level representations, their applications are limited because of their restrictive conditions on missing views. In this section, we introduce LHGN, a more flexible network to address this issue. As shown in Fig.~\ref{framework}, LHGN includes several view-specific encoder networks, a latent heterogeneous graph construction layer, a graph representation learning layer, an aggregation layer, and a classification layer. Detailed information about each part of LHGN is presented in the following subsections.	
	
	\newtheorem{definition}{Definition}
	\begin{definition}
		\textbf{Incomplete Multi-view Classification.}
		Consider a dataset
		$\left\{\mathbf{X}^{(v)}, v=1,2, \ldots, V\right\}$ with $N$ instances, $C$ categories, and $V$ views, where
		$\mathbf{X}^{(v)} \in \mathbf{R}^{d_{v} \times N}$ is the $v$-th view of the dataset and $y_n$ is the class label. Incomplete multi-view classification  aims to design a model and train a classifier using training data in which several samples have missing views, so that a new instance $S_n$ with an arbitrary possible missing-view pattern can be accurately classified \cite{partial}.

	\end{definition}
	
	\subsection{View-specific Encoder Networks}
	
	To deal with samples that have arbitrary missing-view patterns in a flexible manner, we project samples with arbitrary missing-view patterns into a common latent space. The learned latent representation then encodes the observed view information effectively and comprehensively.

	The idea behind this approach is that the expression of the hidden layers can extract common expressions from each individual perspective, but this idea is not
	always feasible in certain data environments. Zhang et al. \cite{cpm} proposed a
	representation based on the hidden space, whereby a potential hidden
	layer representation is obtained through learning, and this is then mapped to the original space to obtain the original data.
	If we denote the implicit space representation of the $n$-th
	sample as $h_n$ and the implicit space mapping encoder as $E_v$, then
	the optimization objective of the elastic implicit space representation is formulated as follows:
	\begin{equation}\label{Lr}
	L_{r}\left(\mathcal{S}_{n}, \mathbf{h}_{n} ; \mathbf{\Theta}_{r}\right)=\sum_{v=1}^{V}
	s_{n
		v}\left\|f_{v}\left(\mathbf{h}_{n} ;
	\mathbf{\Theta}_{r}^{(v)}\right)-\mathbf{x}_{n}^{(v)}\right\|^{2},
	\end{equation}
	where $L_{r}$ is the reconstruction criterion using the common latent representation, which aims to learn the bidirectional mapping between the original data space and the common embedding space. {\small $f_{v}\left(\mathbf{h}_{n} ;
		\mathbf{\Theta}_{r}^{(v)}\right)$} is
	the reconstruction network for the $v$-th view parameterized by
	{\small $\mathbf{\Theta}_{r}^{(v)}$}, and $\mathbf{x}_{n}^{(v)}$ represents the input of view $v$ with $n$ samples.
	$S_n$ is the indicator matrix introduced to identify the present samples from the missing samples for each view. The $v$-th element of this matrix, $s_{n v}$, is defined as follows:
	\begin{equation}
	s_{nv}=\left\{\begin{array}{l}
	1, \text { if the } n \text {-th instance has the } v \text {-th view } \\0, \text { otherwise }
	\end{array}\right. .
	\end{equation}	
	
	To map the hidden space $\mathbf{h}_{n}$ back to the original space $\mathbf{x}_{n}$, the number of encoders should be the same as the number of views. Each encoder is implemented by a multilayer perceptron, and the hidden space $h_n$ is also learnable. In this way, $h_n$ encodes comprehensive information from the different available views, and projects different samples into a common space (regardless of their lost patterns), making them comparable. Ideally, a complete representation will be learned by minimizing Eq. \eqref{Lr}. Because a complete representation encodes information from different views, it should be more versatile than each single view.
	
	Through the latent space representation, we can learn a common space that enables the samples for any missing
	pattern to be reconstructed. This proves that the space has learned the potential elastic
	representation from the observation perspective.
	
\begin{algorithm}[!ht]
	\caption{Algorithm for LHGN}
	\KwIn{Incomplete multi-view dataset $\{\mathcal{S}_{n},y_{n}\}$,
		hyperparameter $\lambda$, and learning rate $\eta$.
		\\
		\textbf{Initialize}: Initialize
		$\left\{\mathbf{h}_{n}\right\}_{n=1}^{N}$, $\left\{\mathbf{\Theta}_{r}^{(v)}\right\}_{v=1}^{V}$ and $\left\{\mathbf{\Theta}_{g}^{(v)}\right\}_{v=1}^{V}$  with random values.
	}
	
	\While{not converged}
	{
		
		\For{$v=1$ to $V$}
		{
			Update $\mathbf{\Theta}_{r}^{(v)}$ with gradient descent:
			$
			\boldsymbol{\Theta}_{r}^{(v)} \leftarrow \mathbf{\Theta}_{r}^{(v)}-\eta \partial \frac{1}{N} \sum_{n=1}^{N} L_{r}\left(\mathcal{S}_{n}, \mathbf{h}_{n} ; \mathbf{\Theta}_{r}\right) / \partial \boldsymbol{\Theta}_{r}^{(v)}
			$
		}
		\For{$n=1$ to $N$}
		{
			Update $\mathbf{h}_{n}$ with gradient descent:
			
			$\mathbf{h}_{n} \leftarrow \mathbf{h}_{n}-\eta \partial
			\frac{1}{N} \sum_{n=1}^{N}\left(L_{r}\left(\mathcal{S}_{n},
			\mathbf{h}_{n}; \mathbf{\Theta}_{r}\right)+\lambda
			L_{c}\left(y_{n}, y, \mathbf{z}_{n}\right)\right) / \partial
			\mathbf{h}_{n}
			$
		}
		\For{$v=1$ to $V$}
		{
			Update $\mathbf{\Theta}_{g}^{(v)}$ with gradient descent:
			
			$\mathbf{\Theta}_{g}^{(v)} \leftarrow \mathbf{\Theta}_{g}^{(v)} -\eta \partial
			\frac{1}{N}\sum_{n=1}^{N}\left(L_{r}\left(\mathcal{S}_{n},
			\mathbf{h}_{n}; \mathbf{\Theta}_{r}\right)+\lambda
			L_{c}\left(y_{n}, y, \mathbf{z}_{n}\right)\right) / \partial
			\mathbf{\Theta}_{g}^{(v)}
			$
		}	
	}
	\textbf{Output:} Network parameters $\left\{\boldsymbol{\Theta}_{r}^{(v)}\right\}_{v=1}^{V}$, $\left\{\mathbf{\Theta}_{g}^{(v)}\right\}_{v=1}^{V}$ and learned representation $\left\{\mathbf{z}_{n}\right\}_{n=1}^{N}$.
	
	\label{algorithm of lhgn}
\end{algorithm}	
	\subsection{Heterogeneous Graph Construction}
	To better depict the complex relationship between the samples in the common latent
	space, we model the latent layer with missing views using a heterogeneous graph.
	
	\textbf{Neighborhood Constraint:}
	Considering that similar instances should be close in the learned latent space, we propose the neighborhood constraint on the common latent representation of all views to preserve the local structure of data samples.
	Generally, the neighborhood structure can be obtained from a Gaussian-based kernel matrix. We denote the matrices as $A_n$, and the entities in the matrix indicate the similarity between two data samples under a specific view. The detailed formulation is as follows:
	\begin{equation}
	\centering
	{A_n}_{i j}=\left\{\begin{array}{l}
	\exp \left(-\frac{\left\|h_{i}-h_{j}\right\|^{2}}{2 \sigma^{2}}\right), h_{i} \in N_{k}\left(h_{j}\right) \text { or } h_{j} \in N_{k}\left(h_{i}\right) \\
	0, \text { otherwise }
	\end{array}\right.
	,
	\end{equation}	
	where $N_{k}\left(h_{i}\right)$ indicates samples of the $K$ nearest neighbors of $h_{i}$.
	
	\textbf{View-existence Constraint:}
	
	\begin{equation}
	\centering	
	{A_p}_{i j}^{(v)}=\left\{\begin{array}{ll}1,     \text{ if }   \mathbf{x}_{i}^{(v)} \text{ and }  \mathbf{x}_{j}^{(v)} \text{ have same semantics}
	\\
	0,\text{ otherwise } \end{array}\right. .
	\end{equation}	
	Using the inter-view and intra-view similarities, we define the overall similarity matrix as:
	\begin{equation}
	A_p=\left[\begin{array}{ccccc}
	A_p^{(1)},&A_p^{(2)},&A_p^{(3)},&\text{...},&  A_p^{(V)}
	\end{array}\right],
	\end{equation}	
	
	\begin{equation}
	\centering	
	\centering
	A_{v}=A_{n} \cdot A_{p}. \label{hetergenous graph}
	\end{equation}
	Then, we obtain:
	\begin{equation}\label{A}
	A=\left[\begin{array}{ccccc}
	A_{n}\cdot A_p^{(1)},&A_{n} \cdot A_p^{(2)},&\text{...},&  A_{n} \cdot A_p^{(V)}
	\end{array}\right].
	\end{equation}

	\subsection{Graph Representation Learning}
	The heterogeneous graph constructed by Eq. \eqref{A}
	contains different types of graph structures. With these graph structures, we can determine the importance of each node to its neighbors in the latent space, and further aggregate the characteristics of these important neighbors to obtain latent space information with some structural expression. To better capture
	feature information from different types of neighbors, we apply GATs \cite{gat} to aggregate the embedded content of heterogeneous neighbors for each node. The target of each GAT is to obtain sufficient expression power to transfer the input features into high-level output features. GATs take a set of node features as input (for LHGN, the input is the latent representation $\mathbf{h}_{v}=$ $\left\{h_{1}, h_{2}, \ldots, h_{N}\right\}$, where $\mathbf{h}_{v} \in\mathbf{R}^{D},N$ is the total number of current samples; $D$ is the dimension of latent space), and output the adjacency matrix;  $A$ generated by multiple graph learners based on different views. It is then possible to generate $V$ groups of new features  $\mathbf{h}_{v}^{\prime}=\left\{h_{1}^{\prime},
	h_{2}^{\prime}, \ldots, h_{N}^{\prime}\right\}$, where $\mathbf{h}_{v}^{\prime} \in
	\mathbf{R}^{D^{\prime}}$.
	
	First, the GAT transforms the latent representation into a vector suitable for graph semantics, which requires a mapping layer composed of learnable parameters $\mathbf{\Theta}_{g}$, according to:
	
	\begin{equation}
	h_{i}^{\prime^{(v)}}=GAT\left(H_{l}^{(v)}, A^{(v)} ; \mathbf{\Theta}_{g}^{(v)}\right),
	\end{equation}
	where $H_{l}^{(v)}$ is the stacked state of all nodes at layer $l$, $A^{(v)}$ is the graph adjacency matrix, and $\mathbf{\Theta}_{g}^{(v)}$ is the parameter set of the GAT.
	
	After obtaining the new representation $h_i$, the network uses the self-attention mechanism to calculate the importance between nodes. The importance of each two hidden space nodes can be calculated by
	$e_{ij}=a\left(\mathbf{W} h_{i}, \mathbf{W} h_{j}\right)$. $\mathbf{W}$ is a weight matrix, $\mathbf{W} \in \mathbf{R}^{D^{\prime} \times D}$. In this process, the
	network actually calculates the attention mechanism for each two nodes and
	ignores the input graph structure  $A_{v}$. Therefore, during the training process, we
	can
	use an attention mask to inject the graph structure into the
	calculation,
	and only consider nodes $i, j$ that have relations in  $A_{v}$. We compute $e_{ij}$ for nodes $j \in \mathcal{N}_{i}$, where $\mathcal{N}_{i}$ is some neighborhood of node $i$ in the graph. To describe the relationship between nodes, the network
	uses
	the softmax operator to calculate the probability:
	\begin{equation}
	\alpha_{i j}=\operatorname{softmax}_{j}\left(e_{i j}\right)=\frac{\exp
		\left(e_{i j}\right)}{\sum_{k \in \mathcal{N}_{i}} \exp \left(e_{i
			k}\right)}.
	\end{equation}
	The attention mechanism adopts a single-layer feedforward network, and uses
	nonlinear LeakReLU elements. Therefore, the attention
	calculation can
	be converted to:
	\begin{equation}
	\centering
	\alpha_{i j}=\frac{\exp \left(\text { \small{LeakyReLU}
		}\left({\mathbf{a}}^{T}\left[\mathbf{W} h_{i} \|
		\mathbf{W}
		h_{j}\right]\right)\right)}{\sum_{k \in \mathcal{N}_{i}} \exp
		\left(\text {
			\small{LeakyReLU}
		}\left({\mathbf{a}}^{T}\left[\mathbf{W} h_{i}
		\|\mathbf{W} h_{k}\right]\right)\right)},
	\end{equation}
where $||$ is the concatenation operation. Furthermore, to stabilize the learning process, the graph attention mechanism introduces a multihead attention operation in which multiple heads are used to predict the attention, and each head captures different information. This can be considered as a form of multiperspective learning at the attention level. If $K$ heads 	are used to capture attention when joining the network, the graph features of
the end of the fork can be expressed as follows:
	\begin{equation}
	h_{i}^{\prime}=\|_{k=1}^{K} \sigma\left(\sum_{j \in N_{i}} \alpha_{i
		j}^{k}
	\mathbf{W}^{k} h_{j}\right),
	\end{equation}
	where $\mathbf{W}^{k}$ is the corresponding input linear transformation’s weight matrix and $||$ represents concatenation.
	
	Once the graph attention mechanism has obtained the
	graph features of the corresponding $V$ perspectives, the network attempts
	to
	aggregate the graph features from different perspectives. If the graph
	features
	are directly averaged, some information will be lost. Therefore, the
	network
	tries to connect the graph features first, before determining the final
	features
	through the aggregation layer.
	\begin{equation}
	\mathbf{z}_{n}=\mathbf{W}_{a g g}\left(\|_{v=1}^{V}\left(h_{1}^{\prime},
	h_{V}^{\prime}\right)\right),
	\end{equation}
	where $\mathbf{W}_{a g g}$ is the learnable parameter of the polymer layer.

	\subsection{Classification Using Aggregated Representation}
	Zhang \etal{} \cite{cpm}  proposed a training and test consistency strategy in which
	the
	test set is used to fine-tune the hidden expression layer to obtain a more coherent
	hidden
	layer representation. However, this method cannot guarantee
	consistency between the training and test results. In the heterogeneous
	graph
	framework proposed in this paper, training set samples and test set samples can
	interact
	directly in the hidden space, effectively enabling semi-supervised classification by
	transpose learning. In this way, training and test are carried out
	synchronously, which greatly reduces the impact of any inconsistencies.
	Therefore,
	the heterogeneous graph representation network based on the hidden space
	transforms
	the classification problem without multiple perspectives into a semi-supervised
	classification problem in which the training set contains labeled data and the
	test
	set consists of unmarked data. Formally, we denote the classification loss as $\Delta$. For the loss function, the network
	attempts to minimize the cross-entropy of marked nodes between the true
	value and the predicted value.
	\begin{equation}
	\Delta\left(y_{n}, y\right)=\Delta\left(y_{n}, g\left(\mathbf{z}_{n} ; \mathbf{\Theta}_{c}\right)\right),
	\end{equation}
	\begin{equation}
	g\left(\mathbf{z}_{n} ; \boldsymbol{\Theta}_{c}\right)=\arg \max _{y \in \mathcal{Y}} \mathbb{E}_{\mathbf{z} \sim \mathcal{T}(y)} F\left(\mathbf{z}, \mathbf{z}_{n}\right),
	\end{equation}
	where $F\left(\mathbf{z}, \mathbf{z}_{n}\right)=\phi\left(\mathbf{z} ; \mathbf{\Theta}_{c}\right)^{T} \phi\left(\mathbf{z}_{n} ; \mathbf{\Theta}_{c}\right)$, in which $\phi\left(\cdot ; \boldsymbol{\Theta}_{c}\right)$ is the
	feature mapping function for $\mathbf{z}$, and $\mathcal{T}(y)$ is the set of latent representations from class $y$. In our implementation, we set $\phi\left(\mathbf{z} ; \mathbf{\Theta}_{c}\right)=\mathbf{z}$ for simplicity and effectiveness. By considering classification and representation learning together, the misclassification loss is specified as:
	\begin{equation}
	\begin{split}
	L_{c}\left(y_{n}, y, \mathbf{z}_{n}\right)=\max \left(0, \Delta\left(y_{n}, y\right)+\mathbb{E}_{\mathbf{z} \sim \mathcal{T}(y)} F\left(\mathbf{z}, \mathbf{z}_{n}\right)\right.
	\\
	\left.\quad-\mathbb{E}_{\mathbf{z} \sim \mathcal{T}\left(y_{n}\right)} F\left(\mathbf{z}, \mathbf{z}_{n}\right)\right),
	\end{split}
	\end{equation}
	where $
	{y}_{n} $ represents the labeled samples in the training set. Through this
	method, the optimized network can be obtained by backpropagation, and the
	category of the test samples can be determined.
	\subsection{Overall Objective Function}
	By synthesizing the above objectives, the overall optimization problem of LHGN is formulated as:
	\begin{equation}
	\min _{\left\{\mathbf{h}_{n}\right\}_{n=1}^{N}, \mathbf{\Theta}_{r}} \frac{1}{N} \sum_{n=1}^{N} L_{r}\left(\mathcal{S}_{n}, \mathbf{h}_{n} ; \mathbf{\Theta}_{r}\right)+\lambda L_{c}\left(y_{n}, y, \mathbf{z}_{n}\right),
	\end{equation}
	where $L_{r}$ is the reconstruction loss defined in Eq. (\ref{Lr}) and
	$\lambda$ is the weight of the classification loss.

	\section{Experiments}

	\begin{table*}[!ht]
		\centering
		\small
		\caption{Performance comparison under different PERs on four
			datasets (mean$\pm$standard deviation). Higher values indicate better performance. The best results are in {\color{red}red}, and the second best results are in {\color{blue}blue}.}
		\label{table:small dataset}
		\setlength{\tabcolsep}{2.8mm}{
			\begin{tabular}{|c|c|c|c|c|c|c|c|}
				\hline
				Datasets & Methods& 0.0 & 0.1 & 0.2 & 0.3 &0.4 &0.5\\
				\hline
				\multirow{8}*{ORL}
                &FeatCon   & 96.40$\pm$1.13 & 90.95$\pm$2.62  & 83.30$\pm$3.27 & 75.70$\pm$3.15  & 69.28$\pm$2.41  & 70.01$\pm$4.55 \\
                 &CCA\cite{cca}       & 96.65$\pm$1.41 & 75.54$\pm$1.06  & 65.43$\pm$8.04 & 55.31$\pm$5.41  & 45.34$\pm$3.60  & 32.79$\pm$7.32 \\
                 &DCCA\cite{dcca}      & 87.88$\pm$5.07 & 62.00$\pm$4.54  & 52.16$\pm$4.55 & 36.97$\pm$6.64  & 31.67$\pm$8.86 & 20.78$\pm$3.43 \\
                 &DCCAE\cite{dccae}     & 88.90$\pm$4.38 & 58.20$\pm$6.46 & 56.82$\pm$4.62 & 35.47$\pm$6.89 & 29.36$\pm$3.14  & 20.12$\pm$4.18 \\
                 &DMF\cite{dmf}       & 95.87$\pm$1.15 & 90.56$\pm$4.54  & 77.80$\pm$5.58 & 67.20$\pm$5.57  & 56.77$\pm$7.21  & 44.49$\pm$4.07 \\
                 &MDcR\cite{mdcr}      & 96.51$\pm$1.85 & 84.67$\pm$2.07  & 76.63$\pm$4.05 & 71.86$\pm$6.19  & 63.96$\pm$4.62  & 60.83$\pm$6.20 \\
                 &ITML\cite{itml}      & 97.10$\pm$1.04 & 84.09$\pm$4.08  & 84.01$\pm$2.98 & 72.17$\pm$2.47  & 61.03$\pm$5.81  & 56.81$\pm$3.08 \\
                 &LMNN\cite{lmnn}      & 97.37$\pm$1.25 & 85.65$\pm$3.55  & 83.78$\pm$3.84 & 75.15$\pm$4.74  & 66.34$\pm$3.47  & 68.20$\pm$4.65 \\
                 &CPM\cite{cpm}       & 97.36$\pm$0.89 & {\color{red}\textbf{98.20$\pm$0.60}}  & {\color{blue}\textbf{97.05$\pm$1.56}} & {\color{blue}\textbf{96.20$\pm$0.85}}  & 93.07$\pm$3.97  & 88.91$\pm$1.92 \\
				&UEAF\cite{ueaf}  &
        		95.85$\pm$2.48  & 94.79$\pm$2.41  & 92.15$\pm$3.34  & 90.27$\pm$1.91  & 88.15$\pm$2.37  & 87.85$\pm$2.48 
        		\\
        		&COMPLETER\cite{lin2021completer}  &
        		{\color{blue}\textbf{97.52$\pm$1.76}} & 97.35$\pm$1.28  & 96.24$\pm$1.32  & 95.18$\pm$2.68  & {\color{blue}\textbf{95.26$\pm$2.63}} & {\color{blue}\textbf{92.83$\pm$3.18}}
        		\\
				
				& LHGN & {\color{red}\textbf{97.78$\pm$1.29}} & {\color{blue}\textbf{97.37$\pm$1.22}}& {\color{red}\textbf{97.13$\pm$1.36}} &
				{\color{red}\textbf{96.85$\pm$1.53}} &{\color{red}\textbf{95.77$\pm$2.93}} &{\color{red}\textbf{93.46$\pm$1.09}}\\
				\hline
				\multirow{8}*{YaleB}
				&FeatCon   & 88.84$\pm$1.08 & 78.22$\pm$3.21 & 71.65$\pm$2.37 & 61.40$\pm$3.36 & 64.17$\pm$2.79 & 57.79$\pm$4.00 \\
                 &CCA\cite{cca}       & 96.47$\pm$0.95 & 82.97$\pm$5.96 & 70.95$\pm$3.52 & 61.65$\pm$3.57 & 56.11$\pm$6.27 & 44.93$\pm$1.94 \\
                 &DCCA\cite{dcca}      & 94.83$\pm$3.74 & 68.50$\pm$4.77 & 55.91$\pm$4.61 & 48.46$\pm$7.92 & 41.48$\pm$4.29 & 22.30$\pm$8.15 \\
                 &DCCAE\cite{dccae}     & 94.24$\pm$2.66 & 67.48$\pm$3.18 & 48.87$\pm$4.77 & 39.28$\pm$6.57 & 24.17$\pm$1.93 & 16.72$\pm$2.39 \\
                 &DMF\cite{dmf}       & 84.41$\pm$2.64 & 76.37$\pm$3.48 & 69.17$\pm$1.76 & 57.71$\pm$4.47 & 49.37$\pm$2.68 & 48.47$\pm$4.05 \\
                 &MDcR\cite{mdcr}      & 71.54$\pm$2.74 & 64.67$\pm$1.95 & 58.31$\pm$2.74 & 57.84$\pm$2.36 & 55.77$\pm$4.77 & 47.57$\pm$6.35 \\
                 &ITML\cite{itml}      & 91.35$\pm$1.19 & 86.06$\pm$2.40 & 77.66$\pm$2.31 & 69.72$\pm$4.12 & 55.40$\pm$2.19 & 45.87$\pm$4.20 \\
                 &LMNN\cite{lmnn}      & 97.21$\pm$0.48 & 92.15$\pm$3.32 & 85.04$\pm$3.42 & 72.57$\pm$4.45 & 65.08$\pm$4.72 & 52.17$\pm$2.75 \\
                 &CPM\cite{cpm}       & {\color{red}\textbf{98.65$\pm$0.61}} & {\color{red}\textbf{98.55$\pm$0.63}} & {\color{red}\textbf{98.52$\pm$1.18}} & {\color{blue}\textbf{94.81$\pm$1.13}} & {\color{blue}\textbf{92.35$\pm$1.88}} & 91.02$\pm$1.91 \\
				&UEAF\cite{ueaf}  &
        		{\color{blue}\textbf{97.73$\pm$1.27}} & 95.18$\pm$2.71  & 92.61$\pm$2.69  & 85.37$\pm$2.14  & 83.37$\pm$1.41  & 83.19$\pm$2.73 
        		\\
        		&COMPLETER\cite{lin2021completer}  &
        		96.43$\pm$1.65  & {\color{blue}\textbf{95.79$\pm$1.88}} & {\color{blue}\textbf{94.21$\pm$2.63}} & 93.44$\pm$2.48  & 92.32$\pm$3.89 & {\color{blue}\textbf{91.66$\pm$3.55}}
        		\\
				&LHGN & 
				95.39$\pm$1.98 & 94.13$\pm$2.83 & 93.77$\pm$3.29 &  {\color{red}\textbf{94.90$\pm$1.95}}&
				{\color{red}\textbf{94.22$\pm$2.44}} & {\color{red}\textbf{93.38$\pm$1.52}}
				\\
				\hline
				\multirow{8}*{PIE}
				&FeatCon   & 79.82$\pm$1.35 & 64.19$\pm$3.45 & 49.90$\pm$5.69 & 36.22$\pm$2.52 & 29.31$\pm$3.51 & 23.40$\pm$2.22 \\
                 &CCA\cite{cca}       & 88.49$\pm$1.97 & 75.52$\pm$1.83 & 61.82$\pm$4.89 & 55.40$\pm$2.26 & 47.52$\pm$3.66 & 37.47$\pm$2.15 \\
                 &DCCA\cite{dcca}      & 83.27$\pm$3.37 & 70.68$\pm$2.36 & 65.61$\pm$4.46 & 56.22$\pm$3.19 & 42.21$\pm$2.41 & 34.56$\pm$3.05 \\
                 &DCCAE\cite{dccae}     & 85.23$\pm$4.43 & 77.26$\pm$1.26 & 65.21$\pm$3.36 & 54.50$\pm$6.98 & 44.30$\pm$4.43 & 39.00$\pm$1.39 \\
                 &DMF\cite{dmf}       & 83.33$\pm$2.49 & 72.83$\pm$2.84 & 64.05$\pm$4.85 & 52.38$\pm$3.17 & 40.01$\pm$3.68 & 30.95$\pm$3.68 \\
                 &MDcR\cite{mdcr}      & 82.82$\pm$3.76 & 66.04$\pm$5.49 & 55.35$\pm$6.48 & 39.68$\pm$4.07 & 33.49$\pm$2.44 & 27.70$\pm$1.74 \\
                 &ITML\cite{itml}      & 88.37$\pm$3.57 & 73.52$\pm$4.55 & 60.98$\pm$4.70 & 47.86$\pm$5.90 & 38.06$\pm$6.49 & 20.73$\pm$3.16 \\
                 &LMNN\cite{lmnn}      & {\color{red}\textbf{94.44$\pm$1.45}} & 79.96$\pm$2.30 & 66.67$\pm$5.02 & 61.81$\pm$5.55 & 56.71$\pm$5.47 & 46.91$\pm$3.50 \\
                 &CPM\cite{cpm}       & 90.30$\pm$1.06 & {\color{blue}\textbf{88.78$\pm$2.91}} & 85.21$\pm$3.45 & {\color{red}\textbf{82.33$\pm$2.13}} & 74.24$\pm$3.31 & 61.84$\pm$3.45 \\
				&UEAF\cite{ueaf}  &
        		93.52$\pm$1.29  & 87.67$\pm$1.84  & {\color{blue}\textbf{85.74$\pm$2.71}} & 77.21$\pm$2.38  & 70.14$\pm$4.05  & 63.72$\pm$3.16 
        		\\
        		&COMPLETER\cite{lin2021completer}  &
        		89.74$\pm$1.22  & 88.54$\pm$1.32 & 83.52$\pm$1.57  & 79.04$\pm$1.45 & {\color{blue}\textbf{75.13$\pm$1.39}} & {\color{red}\textbf{66.27$\pm$2.81}}
        		\\
				&LHGN &
				{\color{blue}\textbf{94.25$\pm$2.64}} &
				{\color{red}\textbf{88.83$\pm$1.52}} &
				{\color{red}\textbf{86.53$\pm$2.75}} &
				{\color{blue}\textbf{79.16$\pm$3.08}} &
				{\color{red}\textbf{75.87$\pm$1.22}} & {\color{blue}\textbf{64.82$\pm$3.55}}
				\\
				\hline
				\multirow{8}*{CUB}
				 &FeatCon   & 86.28$\pm$2.94 & 75.97$\pm$1.70 & 72.14$\pm$3.06 & 73.21$\pm$2.22 & 68.69$\pm$5.96 & 69.75$\pm$2.81 \\
                 &CCA\cite{cca}       & 84.26$\pm$2.44 & 70.85$\pm$1.89 & 67.93$\pm$4.89 & 56.63$\pm$4.00 & 49.92$\pm$2.26 & 31.73$\pm$4.44 \\
                 &DCCA\cite{dcca}      & 72.15$\pm$2.81 & 57.70$\pm$4.06 & 49.69$\pm$5.39 & 48.33$\pm$2.05 & 40.14$\pm$3.47 & 31.90$\pm$4.28 \\
                 &DCCAE\cite{dccae}     & 76.14$\pm$1.63 & 58.22$\pm$6.35 & 53.86$\pm$2.76 & 45.03$\pm$3.57 & 40.81$\pm$4.25 & 34.88$\pm$4.69 \\
                 &DMF\cite{dmf}       & 53.47$\pm$5.09 & 55.90$\pm$5.72 & 45.00$\pm$2.50 & 38.15$\pm$4.88 & 35.33$\pm$1.90 & 30.41$\pm$2.71 \\
                 &MDcR\cite{mdcr}      & 85.18$\pm$1.97 & 77.46$\pm$2.74 & 76.84$\pm$4.46 & 70.22$\pm$3.27 & 68.78$\pm$3.32 & 69.20$\pm$2.03 \\
                 &ITML\cite{itml}      & 84.01$\pm$3.00 & 82.63$\pm$3.00 & 77.86$\pm$4.46 & 72.60$\pm$4.07 & 70.41$\pm$2.68 & 69.75$\pm$2.81 \\
                 &LMNN\cite{lmnn}      & 86.28$\pm$0.85 & 81.48$\pm$1.37 & 77.82$\pm$2.95 & 72.79$\pm$4.20 & 70.32$\pm$4.10 & 48.11$\pm$5.61 \\
                 &CPM\cite{cpm}       & 89.48$\pm$3.64 & {\color{red}\textbf{88.38$\pm$1.27}} & {\color{blue}\textbf{84.32$\pm$2.58}} & {\color{blue}\textbf{81.36$\pm$3.59}} & {\color{blue}\textbf{77.13$\pm$3.64}} & {\color{blue}\textbf{76.31$\pm$3.67}} \\
				&UEAF\cite{ueaf}  &
        		90.53$\pm$4.87  & 86.44$\pm$4.21  & 83.96$\pm$3.13  & 74.59$\pm$1.09  & 69.53$\pm$1.52  & 67.25$\pm$1.83 
        		\\
        		&COMPLETER\cite{lin2021completer}  &
        		{\color{red}\textbf{91.15$\pm$1.85}} & {\color{blue}\textbf{87.52$\pm$1.25}} & 84.16$\pm$1.66 & 81.34$\pm$4.51 & 76.56$\pm$4.95 & 75.03$\pm$2.63 
        		\\
				&LHGN &
				{\color{blue}\textbf{91.02$\pm$1.84}} &
				86.50$\pm$1.34 &
				{\color{red}\textbf{84.33$\pm$2.72}} &
				{\color{red}\textbf{81.39$\pm$1.78}} &
				{\color{red}\textbf{77.15$\pm$3.84}} &
				{\color{red}\textbf{76.78$\pm$2.56}}
				\\
				\hline
		\end{tabular}}
	\end{table*}
	
	\begin{table*}[!ht]
	\centering
	\small
	\caption{Performance comparison under different PERs on two datasets (mean$\pm$standard deviation). Higher values indicate better performance. The best results are in {\color{red}red}, and the second best results are in {\color{blue}blue}.}
	\setlength{\tabcolsep}{2.8mm}{
		\begin{tabular}{|c|c|c|c|c|c|c|c|}
		\hline
		Datasets & Methods& 0.0 & 0.1 & 0.2 & 0.3 & 0.4 & 0.5\\
		\hline
		\multirow{8}*{HW}
		&FeatCon   & {\color{red}\textbf{98.15$\pm$0.73}} & {\color{red}\textbf{96.92$\pm$0.52}} & 94.48$\pm$0.96 & 93.49$\pm$0.82 & 91.29$\pm$1.42 & 87.14$\pm$2.18 \\
         &CCA\cite{cca}       & 95.41$\pm$2.12 & 82.55$\pm$2.79 & 76.31$\pm$1.21 & 65.84$\pm$2.87 & 59.44$\pm$1.55 & 51.24$\pm$1.46 \\
         &DCCA\cite{dcca}      & 96.06$\pm$0.09 & 78.81$\pm$3.13 & 68.90$\pm$1.80 & 60.62$\pm$2.16 & 51.03$\pm$2.25 & 37.87$\pm$3.53 \\
         &DCCAE\cite{dccae}     & 96.74$\pm$0.49 & 88.64$\pm$1.23 & 80.47$\pm$0.96 & 70.29$\pm$1.45 & 59.08$\pm$2.27 & 50.00$\pm$1.33 \\
         &DMF\cite{dmf}       & 72.28$\pm$1.21 & 63.40$\pm$2.83 & 59.00$\pm$2.70 & 48.28$\pm$6.54 & 44.77$\pm$2.08 & 37.33$\pm$3.83 \\
         &MDcR\cite{mdcr}      & 97.70$\pm$0.36 & 95.76$\pm$0.76 & 95.25$\pm$1.24 & 92.55$\pm$1.09 & 91.27$\pm$0.81 & 87.10$\pm$1.30 \\
         &ITML\cite{itml}      & 96.84$\pm$0.56 & 90.83$\pm$1.62 & 85.45$\pm$2.90 & 81.39$\pm$2.44 & 74.76$\pm$4.15 & 76.41$\pm$1.38 \\
         &LMNN\cite{lmnn}      & {\color{blue}\textbf{98.13$\pm$0.39}} & 94.18$\pm$1.29 & 91.25$\pm$1.33 & 87.89$\pm$2.45 & 83.15$\pm$2.12 & 81.49$\pm$2.03 \\
         &CPM\cite{cpm}       & 95.01$\pm$0.89 & 94.82$\pm$0.75 & 93.67$\pm$1.21 & 93.57$\pm$1.64 & 92.67$\pm$0.47 & 91.01$\pm$2.43 \\
		&UEAF\cite{ueaf}  &
		93.28$\pm$2.06  & 92.39$\pm$2.23  & 92.81$\pm$1.83  & 92.23$\pm$2.68  & 92.47$\pm$2.35  & 91.57$\pm$2.42 
		\\
		&COMPLETER\cite{lin2021completer}  &
		96.91$\pm$1.89 & 95.68$\pm$1.26  & {\color{blue}\textbf{95.57$\pm$1.65}} & {\color{blue}\textbf{94.78$\pm$1.27}} & {\color{blue}\textbf{93.27$\pm$1.68}} & {\color{blue}\textbf{92.82$\pm$2.15}}
		\\
		&  LHGN  &
		95.97$\pm$1.94 &
		{\color{blue}\textbf{95.79$\pm$1.79}} &
		{\color{red}\textbf{95.64$\pm$1.27}} &
		{\color{red}\textbf{95.31$\pm$1.73}} &
		{\color{red}\textbf{94.18$\pm$2.49}} &
		{\color{red}\textbf{93.05$\pm$1.91}}\\
		\hline
		\multirow{8}*{Animal}
		&FeatCon   & 82.30$\pm$1.30 & 76.32$\pm$0.68 & 71.93$\pm$0.52 & 68.11$\pm$0.82 & 64.87$\pm$0.55 & 60.63$\pm$1.39 \\
         &CCA\cite{cca}       & 47.30$\pm$0.68 & 37.30$\pm$0.82 & 33.28$\pm$0.78 & 28.93$\pm$0.84 & 24.87$\pm$0.54 & 8.53 $\pm$0.41 \\
         &DCCA\cite{dcca}      & 38.33$\pm$0.38 & 7.72 $\pm$0.20 & 6.45 $\pm$0.33 & 5.19 $\pm$0.14 & 4.24 $\pm$0.20 & 2.63 $\pm$0.31 \\
         &DCCAE\cite{dccae}     & 44.50$\pm$0.33 & 27.38$\pm$0.85 & 22.77$\pm$0.69 & 19.19$\pm$1.29 & 15.59$\pm$0.47 & 11.79$\pm$0.42 \\
         &DMF\cite{dmf}       & 79.38$\pm$0.24 & 71.91$\pm$1.60 & 63.46$\pm$1.02 & 55.16$\pm$0.96 & 47.41$\pm$0.60 & 39.93$\pm$0.73 \\
         &MDcR\cite{mdcr}      & 82.17$\pm$0.24 & 76.93$\pm$1.10 & 72.25$\pm$1.52 & 68.65$\pm$0.47 & 66.15$\pm$0.94 & 61.36$\pm$0.79 \\
         &ITML\cite{itml}      & 83.59$\pm$0.68 & 77.19$\pm$1.01 & 70.85$\pm$0.88 & 66.54$\pm$0.74 & 64.73$\pm$0.71 & 61.27$\pm$0.86 \\
         &LMNN\cite{lmnn}      & 83.84$\pm$1.39 & 78.03$\pm$1.08 & 70.85$\pm$0.85 & 66.50$\pm$1.16 & 64.95$\pm$0.43 & 61.01$\pm$1.16 \\
         &CPM\cite{cpm}       & 86.88$\pm$0.68 & 84.13$\pm$0.41 & {\color{blue}\textbf{81.00$\pm$1.18}} & 76.80$\pm$0.45 & 73.31$\pm$0.78 & 67.33$\pm$3.21 \\
		&UEAF\cite{ueaf}  &
		{\color{blue}\textbf{88.67$\pm$1.97}} & 83.17$\pm$2.67  & 78.87$\pm$2.23  & 75.68$\pm$2.19  & 73.76$\pm$2.14  & 70.61$\pm$1.98 
		\\
		&COMPLETER\cite{lin2021completer}  &
		{\color{red}\textbf{88.93$\pm$2.38}} & {\color{blue}\textbf{85.16$\pm$1.73}} & 80.87$\pm$2.33 & {\color{blue}\textbf{77.01$\pm$2.66}} & {\color{blue}\textbf{73.82$\pm$2.85}} & {\color{blue}\textbf{70.64$\pm$2.37}}
		\\
		&   LHGN  &
		88.38$\pm$1.24 &
		{\color{red}\textbf{85.94$\pm$0.95}} &
		{\color{red}\textbf{81.01$\pm$1.30}} &
		{\color{red}\textbf{77.03$\pm$2.29}} &
		{\color{red}\textbf{74.52$\pm$1.61}} &
		{\color{red}\textbf{70.67$\pm$3.60}}
		\\
		\hline
		
		\end{tabular}}
	\label{table:medium dataset}
\end{table*}
	
	\begin{figure*}[h]
		\centering
		\includegraphics[width=1\linewidth]{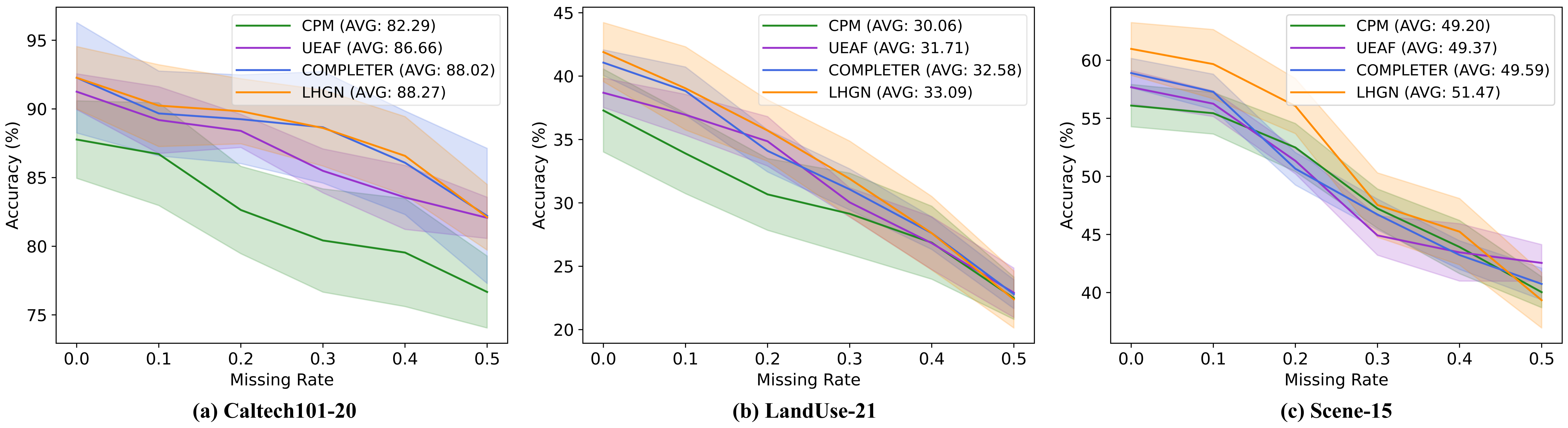}
		\caption{ Performance comparison under different PERs on three datasets.}
		\label{fig:new_datasets}
	\end{figure*}

	\begin{figure*}
		\centering
		\includegraphics[width=0.9\linewidth]{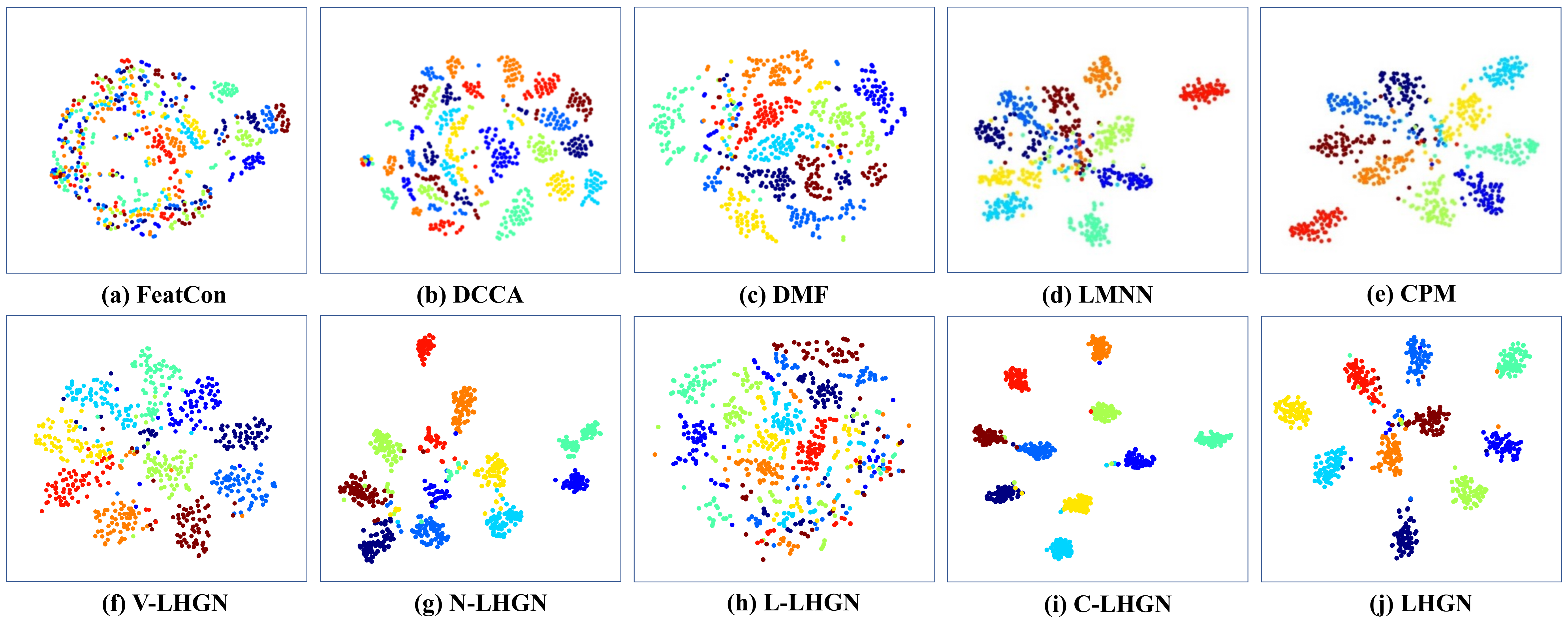}
		\caption{t-SNE visualization of (a) concatenation features,
			and the consensus
			representation obtained by (b) DCCA, (c) DMF, (d) LMNN, (e) CPM, (f) V-LHGN, (g) N-LHGN, (h) L-LHGN, (i) C-LHGN,
			and (j) LHGN on the YaleB dataset with a PER of
			0.5.}
		\label{fig:tsne}
	\end{figure*}
	
		\begin{figure*}
		\centering
		\includegraphics[width=0.8\linewidth]{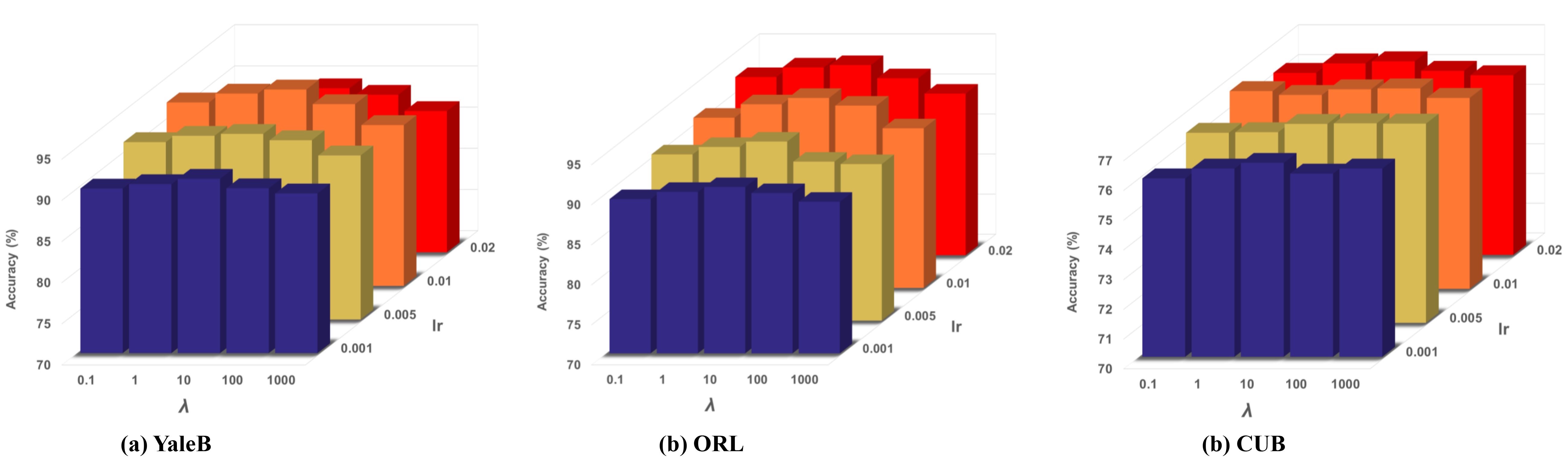}
		\caption{Accuracy (\%) against $\lambda$ and learning rate ($lr$) of LHGN on
			(a) YaleB, (b) ORL, and (c) CUB datasets with PER = 0.5.}
		\label{parameter}
	\end{figure*}
	
			\begin{figure*}[h]
		\centering
		\includegraphics[width=1\linewidth]{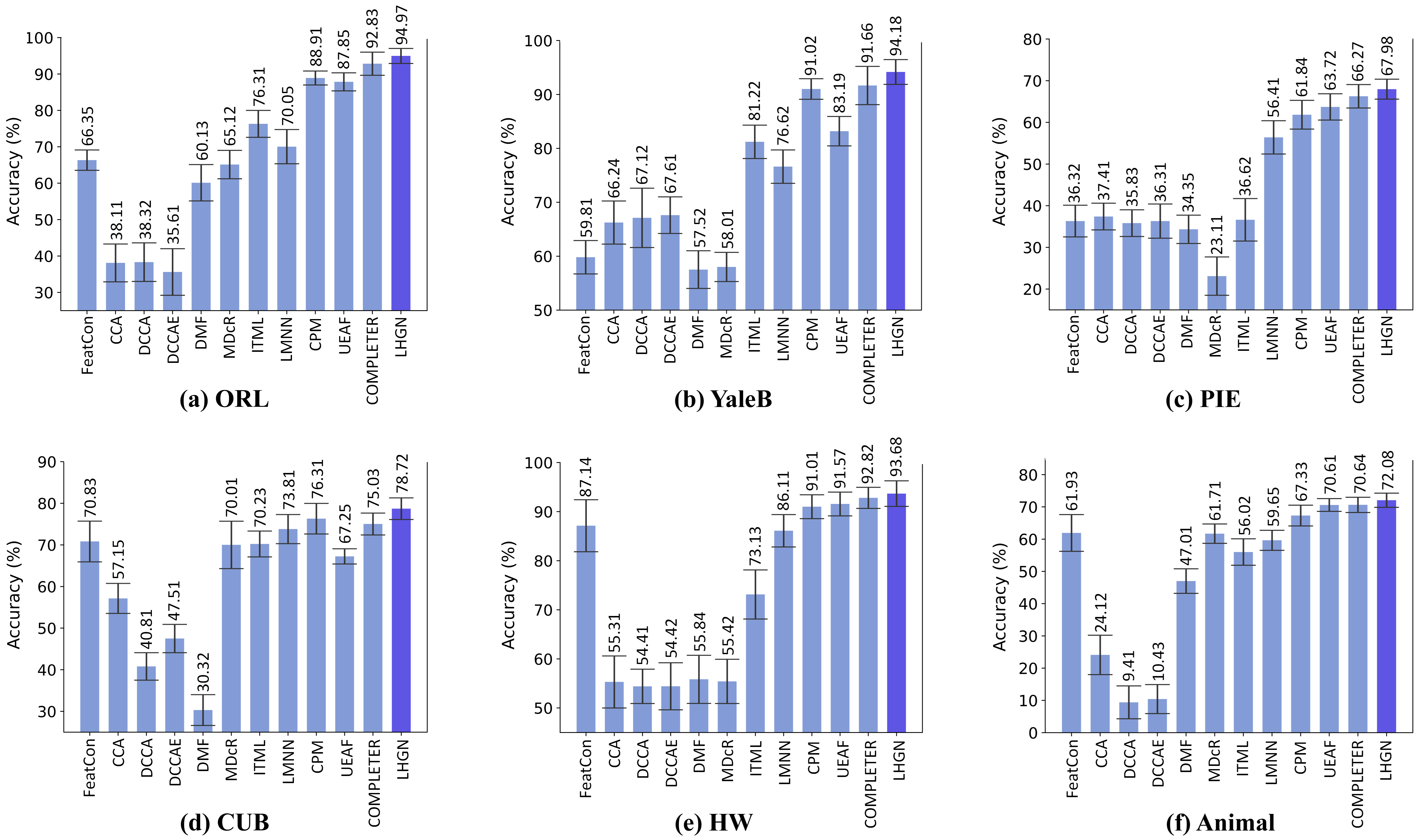}
		\caption{ Performance comparison with view completion using cascaded residual autoencoder (CRA).}
		\label{fig:cra}
	\end{figure*}

		\begin{figure*}
		\centering
		\includegraphics[width=1\linewidth]{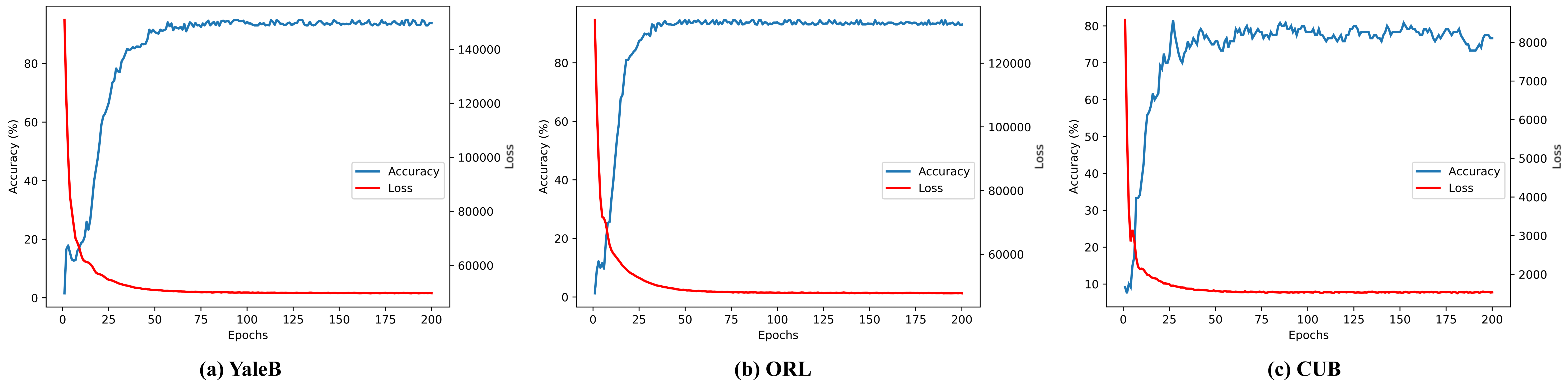}
		\caption{Convergence curves with PER = 0.5 on (a) YaleB dataset, (b) ORL dataset, and (c) CUB dataset.}
		\label{convergence}
	\end{figure*}
	
	\begin{figure*}
		\centering
		\includegraphics[width=1\linewidth]{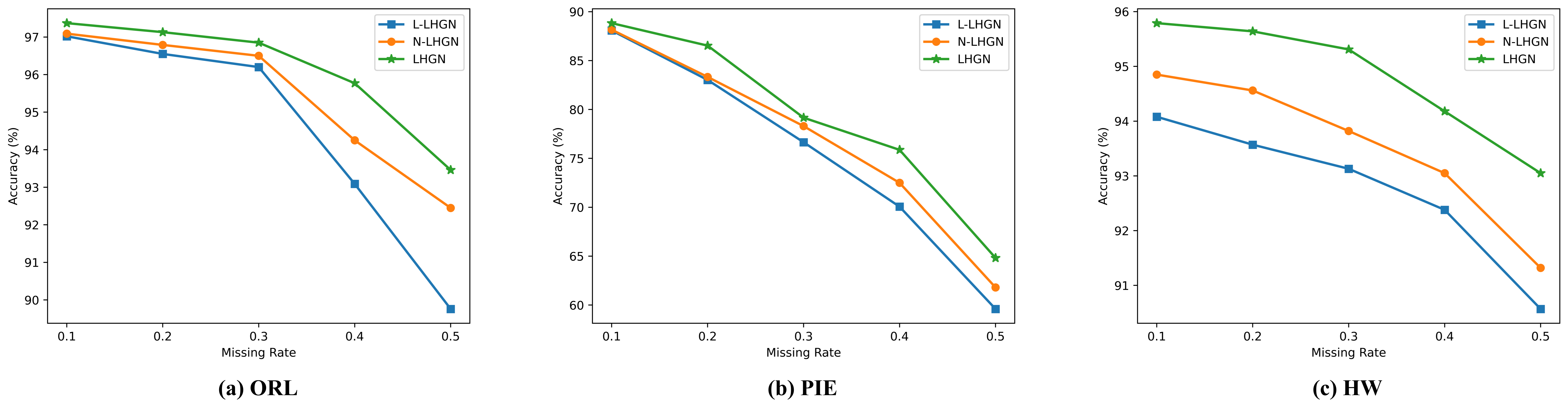}
		\caption{Ablation study using three multi-view datasets.
			Higher values indicate better performance.}
		\label{ablation}
	\end{figure*}
	\subsection{ Datasets}
	
				\begin{table*}[!ht]
	\centering
	\small
	\caption{Runtime (s) comparison on nine datasets.}
	\setlength{\tabcolsep}{2mm}{
		\begin{tabular}{|c|c|c|c|c|c|c|c|c|c|}
		\hline
		Methods & ORL& YaleB & PIE & CUB & HW & Animal & Caltech101-20 & LandUse-21 & Scene-15\\
		\hline
		CPM\cite{cpm}  &30&40&25&25&60&70&60&50&65
		\\
		\hline
        UEAF\cite{ueaf}  &40&45&40&35&135&150&140&130&160
		\\
		\hline
        COMPLETER\cite{lin2021completer}  &35&40&30&30&65&75&55&45&70
		\\
		\hline
        LHGN  &65&70&65&60&255&260&255&245&280
        \\
		\hline
		\end{tabular}}
	\label{table:cost}
\end{table*}
	
	\begin{itemize}
		
		\item ORL\footnote{https://www.cl.cam.ac.uk/research/dtg/attarchive/
		\\
		facedatabase.html}: ORL is a popular face database in the field of face recognition. It contains 400 face images provided by 40 volunteers, with 10 face images from each person. Three types of features, i.e., LBP, Gabor, and intensity, are extracted as the three views for representing every face image.
		
		\item YaleB\footnote{ftp://plucky.cs.yale.edu/CVC/pub/images/yalefacesB/}: Similar to previous work \cite{georghiades2001few}, we choose a subset with 650 samples and categorize it into 10 clusters. Three types of features: intensity, LBP and Gabor are extracted.
		
		\item PIE\footnote{http://www.cs.cmu.edu/afs/cs/project/PIE/MultiPie/Multi-Pie/Home.html}: A subset contains 680 facial images of 68 subjects, and the intensity, LBP, and Gabor features have been extracted.
		
		\item CUB \cite{wah2011caltech}: Caltech-UCSD Birds (CUB) contains 11,788 bird images associated with text descriptions from 200 different categories, and the first 10 categories are used. We extracted features based on images using GoogLeNet and text using doc2vec \cite{le2014distributed}.
		
		\item Handwritten (HW)\footnote{https://archive.ics.uci.edu/ml/datasets/Multiple+Features}: This dataset contains 200 handwritten images of each of the digits 0--9. Six kinds of features, i.e., pixel average, Fourier coefficient, profile correlation, Zernike moment, Karhunen--Loeve coefficient, and morphological, have been extracted from every sample to give six views, where the feature dimensions are 240, 76, 216, 47, 64, and 6, respectively.
		
		\item Animal \cite{22}: There are 10,158 images provided by 50 classes in the dataset, where each image is represented by two kinds of features extracted by DECAF \cite{decaf} and VGG19 \cite{vgg},
		respectively.
		
		\item Caltech101-20~\cite{2015Large}: Caltech101-20 consists of 2,386 images of 20 subjects with the views of HOG and GIST features.
		
		\item LandUse-21~\cite{yang2010bag}: LandUse-21 consists of 2,100 satellite images from 21 categories with PHOG and LBP features.
				
		\item Scene-15~\cite{2005A}: Scene-15 consists of 4,485 images distributed over 15 scene categories, which is with PHOG and GIST features. 
		
	\end{itemize}

	\subsection{ Methods for Comparison}
	
	To assess the effectiveness of the proposed LHGN, we compared its performance against that of the
	following methods:
	
	(1) \textbf{FeatCon} directly concatenates the features of different
	views,
	and then performs the classification task on the concatenated features.
	
	(2) \textbf{CCA} \cite{cca} maps multiple types of features
	to a common space, and then concatenates the low-dimensional features of
	different views.
	
	(3) \textbf{DCCA} \cite{dcca} uses a neural network to learn low-dimensional features and then
	concatenates them.
	
	(4) \textbf{DCCAE} \cite{dccae}
	uses autoencoders to acquire a common representation, and then combines
	the low-dimensional features of these projections.
	
	(5) \textbf{DMF-MVC} (Deep Semi-NMF for Multi-View Clustering)
	\cite{dmf} uses nonnegative matrix factorization and a deep network
	structure to obtain a common feature representation.
	
	(6) \textbf{MDcR} (Multi-view Dimension co-Reduction) \cite{mdcr} uses kernel matching to regularize the dependencies among multiple views, and projects each view into a low-dimensional space.
	
	(7) \textbf{ITML} (Information Theoretic Metric Learning) \cite{itml}
	uses the Mahalanobis distance function as a metric, and
	transforms this into a specific Bregman optimization problem.
	
	(8) \textbf{LMNN} (Large Margin Nearest Neighbors) \cite{lmnn} searches
	the Mahalanobis distance metric to optimize a $k$-nearest neighbor (KNN) classifier. For
	metric learning method, the original features of multiple views are
	connected, and then the projection from the learned metric
	matrix is used to obtain a new representation.
	
	(9) \textbf{CPM-Nets} (Cross Partial Multi-View Networks) \cite{cpm} uses a degradation process that mimics
	data transmission to learn the latent multi-view representation, allowing
	the optimal trade-off between consistency and complementarity across
	different views to be achieved.
		
	(10) \textbf{UEAF} \cite{ueaf} proposes a Unified Embedding Alignment Framework (UEAF) for robust incomplete multi-view clustering. In particular, a locality-preserved reconstruction term is introduced to infer the missing views such that all views can be naturally aligned. A consensus graph is adaptively learned and embedded via the reverse graph regularization to guarantee the common local structure of multiple views and in turn can further align the incomplete views and inferred views. Moreover, an adaptive weighting strategy is designed to capture the importance of different views.
	
	(11) \textbf{COMPLETER} (inCOMPlete muLti-view clustEring via
conTrastivE pRediction) \cite{lin2021completer} provides a theoretical framework that unifies the consistent representation learning and cross-view data recovery. To be specific, the informative and consistent representation is learned by maximizing the mutual information across different views through contrastive learning, and the missing views are recovered by minimizing the conditional entropy of different views through dual prediction.
	
	\subsection{ Experimental Setup }
	
	\textbf{Incomplete Multi-view Construction.}  To generate incomplete multi-view datasets from complete multi-view datasets, all baselines were randomly selected partial examples/instances under different partial
	example ratios (PERs). The PER is defined as $\beta=\frac{\sum_{v} M_{v}}{V \times
		N}$, where $M_{v}$ indicates the number of samples without the $v$-th view.
	Specifically, in constructing such incomplete data, views are removed at random from each instance until the desired PER is attained, under the condition that all samples retain at least one view.
	
	In our experiment, each dataset was randomly divided into a training
	set (80\%)
	and a test set (20\%).
	For all methods, we tuned the parameters through a 10-fold cross-validation process. To remove any bias introduced by the dataset construction
	process, we ran each algorithm 10 times and report the average values of
	the
	performance measures.
	
	\subsection{  Experimental Results and Analysis }
	We first evaluate the performance of the proposed algorithm using six multi-view
	datasets (Table \ref{table:small dataset} and Table \ref{table:medium dataset}). The LHGN model is then evaluated by comparing it with state-of-the-art multi-view representation learning methods. The parameters in the comparative methods are set to the default values in the original papers. In the experiments, the PER ranged from 0.1--0.5 at intervals of 0.1. Experiments were also conducted with PER = 0, i.e., all the views are complete.
		
	The average accuracy and standard deviation are presented in Tables~\ref{table:small dataset} and~\ref{table:medium dataset}. From the results, we can make
	the following observations:
	\begin{itemize}
		\item The results clearly show that LHGN has the advantage in the case of missing views, which indicates that the proposed method would be highly useful in handling multi-view classification tasks with multiple missing data, e.g., the scenarios where PER is high.
		
		\item In general, as the incomplete sample ratio increases, the performance of all methods decreases. This demonstrates that the absence of views is harmful to multi-view learning.
		\item Our model is rather robust to missing views. Generally, as the PER increases, the classification performance of all approaches drops. However, the performance of LHGN does not drop as much as that of other methods. For example, as the missing view rate increases from PER = 0.0 to PER = 0.5, the performance decline with the ORL dataset is less than 5\%, and the decline is less than 3\% on the YaleB and HW datasets. This implies that our method can effectively explore
		the complex relationship between views and samples, even with a relatively large incomplete sample ratio.
	\end{itemize}
	
	To better evaluate LHGN, we also conducted experiments on three datasets, Caltech101-20, LandUse-21, and Scene-15. The comparison method includes three recent SOTA models, CPM, UEAF, and COMPLETER. The results show that the proposed method can outperform other methods in Fig. \ref{fig:new_datasets}, which demonstrates its effectiveness.

		

	In Fig.~\ref{fig:tsne}, we used the t-SNE \cite{tsne} to visualize the features or consensus
	representations obtained by different methods on the YaleB dataset with a PER of 0.5. The consensus representation
	obtained by the proposed LHGN exhibits the best separability for different classes, where samples from the same class are naturally gathered together and the gap between different groups is obvious. This demonstrates the effectiveness of LHGN for IMVL. 	
	
	In order to further explore the impact of completion view on model performance, we visualized the model performance after completion through CRA \cite{CRA}. We filled the missing views with the imputation method in CRA and include a performance comparison in Fig.~\ref{fig:tsne}. $C-LHGN$ denotes that LHGN with view completion using CRA. Since CRA needs a subset of samples with complete views in training, we set 50$\%$ of data as complete-view samples and the remaining are samples with missing views (PER = 0.5). Fig.~\ref{fig:tsne} (i) and (j) distinctly show that LHGN filled with CRA by using part of samples with complete views is more compact and the margins between different classes become more clear due to capturing the correlation of different views. We can see that with proper data filling methods, the performance of our method can be further improved, which demonstrates its effectiveness and flexibility. Moreover, this also infers that the proposed method can handle data with missing views effectively without using additional filling techniques. In addition, as shown in Fig.~\ref{fig:cra}, by comparing the effects of CRA completion view, our method has more potential than some methods of built-in completion view.
	
	Despite the effectiveness, a natural concern is that the computational cost of graphs is extremely high. We explore this point by comparing the runtime of SOTA models, CPM, UEAF, COMPLETER, and LHGN, respectively on nine datasets. We implement our experiment with an NVIDIA 3090 GPU. The results are shown in Table~\ref{table:cost}. It can be seen that LHGN requires more computational cost than other models. It is worth noting that the runtime of LHGN is less than two times than that of UEAF. Considering the effectiveness and flexibility of the proposed method, such additional cost is acceptable.

	\subsection{Parameter Analysis}
	There are three main parameters in our proposed model: learning rate $lr$, weight parameter $\lambda$, and the number of attention heads $K$. In this section, we investigate how the performance varies with changes to these two parameters. We set the PER to 0.5, and selected $\lambda$ from $\{0.1, 1, 10, 100, 1000\}$ and $lr$ from $\{0.001, 0.005, 0.01, 0.02\}$ in experiments using the YaleB, ORL, and CUB datasets. Fig.~\ref{parameter} shows the accuracy with respect to $\lambda$ and $lr$. From
	Fig.~\ref{parameter} we can see that LHGN achieves consistently good performance when $\lambda$ is around 10 and $lr$ is around 0.01. Based on these results, we set $\lambda$ = 10 and $lr$ = 0.01 in subsequent experiments. 
	
\begin{figure}
    \centering
    \includegraphics[width=0.4\textwidth]{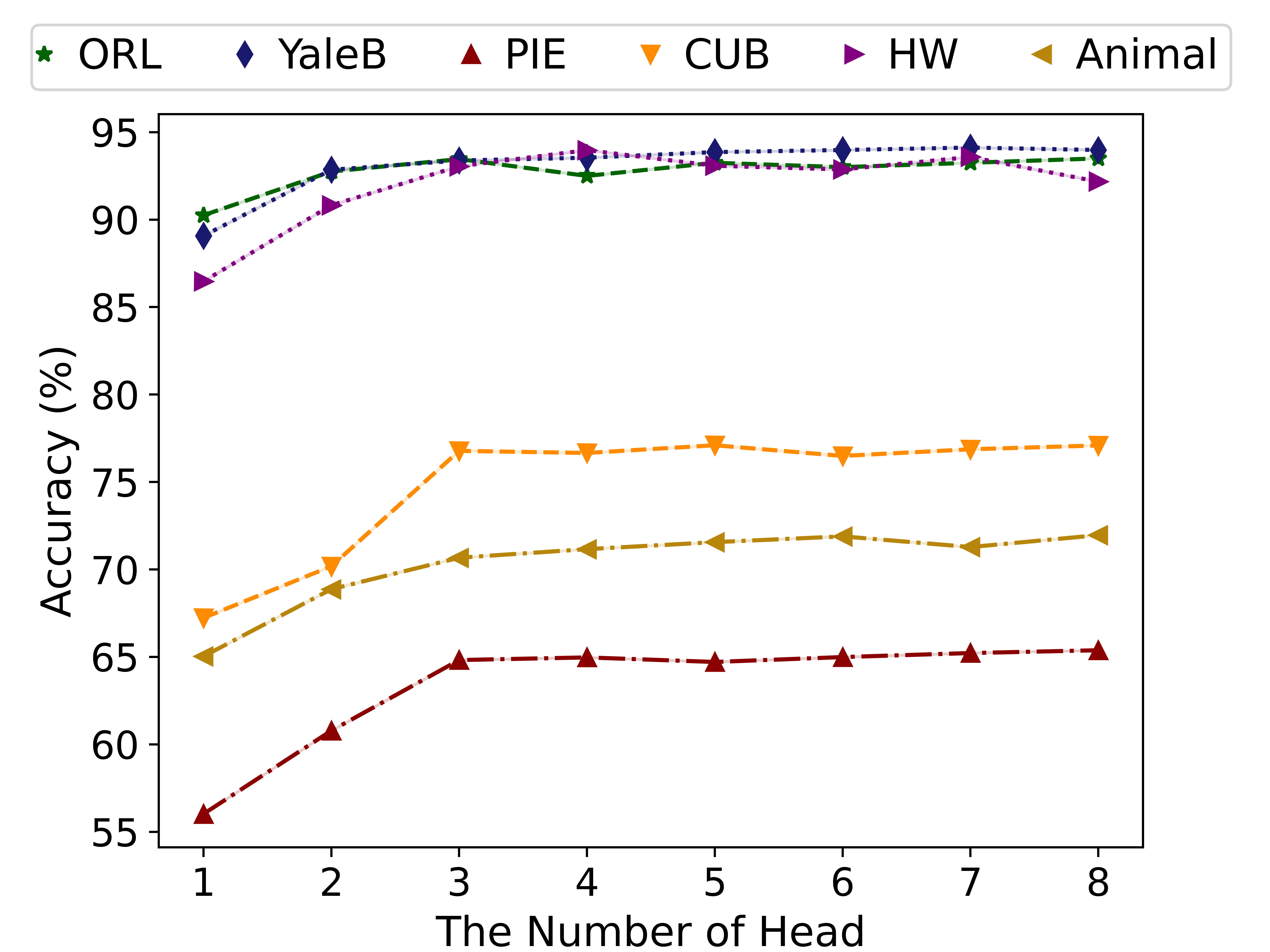}
    \caption{Multihead attention analysis using six multi-view datasets with PER = 0.5.}\label{fig:k_head}
    \end{figure}

    In terms of $K$, it is a hyperparameter which is determined according to the specific downstream task. We compared and analyzed the determination of $K$ value on six different datasets, as shown in Fig.~\ref{fig:k_head}.
    In general, a larger value of $K$ helps to improve classification performance, but incurs high computational cost due to the use of too many heads.

	\subsection{Convergence Study}
	To prove the effectiveness of the employed optimization
	strategy for the objective function of our proposed LHGN, the relationship between the loss of the LHGN and the classification performance on the YaleB, ORL, and CUB datasets with PER = 0.5 is shown in Fig.~\ref{convergence}. In each subfigure, the x-axis denotes the number of epochs, while the left and right y-axes denote the classification performance and corresponding loss value, respectively. LHGN converges quickly for all datasets. Specifically, it converges within 75 epochs on the YaleB, ORL, and CUB datasets. This indicates that our proposed LHGN method efficiently obtains an optimized solution.

	\subsection{Ablation Study}

To explore the importance of neighborhood and view-existence constraints, we study the contribution of having neighborhood and view-existence constraints via t-SNE visualization, as shown in Fig.~\ref{fig:tsne}. $V-LHGN$, $N-LHGN$, and $L-LHGN$ denote LHGN with view-existence constraint, neighborhood constraint, and using only latent representation, respectively. It can be seen from subfigures (f) - (j) that: (1) 
Either neighborhood or view-existence constraint can benefit the learning in the latent space. (2) LHGN with both constraints is more compact and the margins between different classes become more clear compared to using one of the constraints. Besides qualitative analysis, we conducted a quantitative ablation study of both constraints. 

To further verify the contribution of each module in LHGN, we performed an ablation study using the ORL, PIE, and HW datasets. The aim of this experiment is to isolate the effects of the neighborhood constraint and the view-existence constraint. The results are shown in Fig.~\ref{ablation}. $L-LHGN$ uses the learned latent representation for classification, while 
$N-LHGN$ uses the neighborhood constraint to construct a graph on the learned latent representation followed by graph attention for feature aggregation. As we can see from Fig.~\ref{ablation}, $N-LHGN$ performs better than $L-LHGN$, which demonstrates the effectiveness of the neighborhood constraint and feature learning on the graph. Furthermore, LHGN using both constraints performs better than the other two methods, demonstrating the effectiveness of the view-existence constraint.
	
\section{Conclusion }
	In this paper, we proposed a novel LHGN model that provides sufficient flexibility for arbitrary missing-view patterns.  We developed a neighborhood constraint and a view-existence constraint, which allow us to construct a heterogeneous graph in the common latent space. Then, graph learning technique is used to aggregate the features in the constructed graph. In this way, the learned latent space can capture the complex and flexible relationship between samples and views. In addition, by conducting transductive learning on the graph, we not only enhance the interaction between samples, but also avoid any inconsistencies between training and test phases. The superior performance of LHGN has been validated on several incomplete multi-view datasets and in a comparison with many state-of-the-art IMVL methods. In the future, we can extend the study to various scenarios, such as fine-grained recognition~\cite{9115215} and wireless communication~\cite{9326401}. 
	
	\bibliographystyle{IEEEtran}
	\bibliography{egbib}

\begin{thebibliography}{10}
\providecommand{\url}[1]{#1}
\csname url@samestyle\endcsname
\providecommand{\newblock}{\relax}
\providecommand{\bibinfo}[2]{#2}
\providecommand{\BIBentrySTDinterwordspacing}{\spaceskip=0pt\relax}
\providecommand{\BIBentryALTinterwordstretchfactor}{4}
\providecommand{\BIBentryALTinterwordspacing}{\spaceskip=\fontdimen2\font plus
\BIBentryALTinterwordstretchfactor\fontdimen3\font minus
  \fontdimen4\font\relax}
\providecommand{\BIBforeignlanguage}[2]{{%
\expandafter\ifx\csname l@#1\endcsname\relax
\typeout{** WARNING: IEEEtran.bst: No hyphenation pattern has been}%
\typeout{** loaded for the language `#1'. Using the pattern for}%
\typeout{** the default language instead.}%
\else
\language=\csname l@#1\endcsname
\fi
#2}}
\providecommand{\BIBdecl}{\relax}
\BIBdecl

\bibitem{2019Jointly}
Y.~Chen, X.~Xiao, and Y.~Zhou, ``Jointly learning kernel representation tensor
  and affinity matrix for multi-view clustering,'' \emph{IEEE Transactions on
  Multimedia}, vol.~PP, no.~99, pp. 1--1, 2019.

\bibitem{2015Multi}
S.~K. Kuanar, K.~B. Ranga, and A.~S. Chowdhury, ``Multi-view video
  summarization using bipartite matching constrained optimum-path forest
  clustering,'' \emph{IEEE Transactions on Multimedia}, vol.~17, no.~8, pp.
  1166--1173, 2015.

\bibitem{OlivaT01}
A.~Oliva and A.~Torralba, ``Modeling the shape of the scene: {A} holistic
  representation of the spatial envelope,'' \emph{Int. J. Comput. Vis.},
  vol.~42, no.~3, pp. 145--175, 2001.

\bibitem{LadesVBLMWK93}
M.~Lades, J.~C. Vorbr{\"{u}}ggen, J.~M. Buhmann, J.~Lange, C.~von~der Malsburg,
  R.~P. W{\"{u}}rtz, and W.~Konen, ``Distortion invariant object recognition in
  the dynamic link architecture,'' \emph{{IEEE} Trans. Computers}, vol.~42,
  no.~3, pp. 300--311, 1993.

\bibitem{hog}
D.~G. Lowe, ``Distinctive image features from scale-invariant keypoints,''
  \emph{International Journal of Computer Vision}, vol.~60, no.~2, pp. 91--110,
  2004.

\bibitem{sift}
T.~Ojala, M.~Pietik{\"{a}}inen, and T.~M{\"{a}}enp{\"{a}}{\"{a}}, ``Gray scale
  and rotation invariant texture classification with local binary patterns,''
  in \emph{Computer Vision - {ECCV} 2000, 6th European Conference on Computer
  Vision, Dublin, Ireland, June 26 - July 1, 2000, Proceedings, Part {I}},
  2000, pp. 404--420.

\bibitem{ZhaoDF17}
H.~Zhao, Z.~Ding, and Y.~Fu, ``Multi-view clustering via deep matrix
  factorization,'' in \emph{Proceedings of the Thirty-First {AAAI} Conference
  on Artificial Intelligence, February 4-9, 2017, San Francisco, California,
  {USA}}, 2017, pp. 2921--2927.

\bibitem{Sun13}
S.~Sun, ``A survey of multi-view machine learning,'' \emph{Neural Comput.
  Appl.}, vol.~23, no. 7-8, pp. 2031--2038, 2013.

\bibitem{Ding-et-al:2016}
Z.~Ding and Y.~Fu, ``Robust multi-view subspace learning through dual low-rank
  decompositions,'' in \emph{Proceedings of the Thirtieth {AAAI} Conference on
  Artificial Intelligence, February 12-17, 2016, Phoenix, Arizona, {USA}},
  2016, pp. 1181--1187.

\bibitem{xu:mvl}
C.~Xu, D.~Tao, and C.~Xu, ``Multi-view learning with incomplete views,''
  \emph{IEEE Trans. Image Process.}, vol.~24, no.~12, pp. 5812--5825, 2015.

\bibitem{NieCLL18}
F.~Nie, G.~Cai, J.~Li, and X.~Li, ``Auto-weighted multi-view learning for image
  clustering and semi-supervised classification,'' \emph{{IEEE} Trans. Image
  Process.}, vol.~27, no.~3, pp. 1501--1511, 2018.

\bibitem{8587193}
C.~Tang, X.~Zhu, X.~Liu, M.~Li, P.~Wang, C.~Zhang, and L.~Wang, ``Learning a
  joint affinity graph for multiview subspace clustering,'' \emph{IEEE
  Transactions on Multimedia}, vol.~21, no.~7, pp. 1724--1736, 2019.

\bibitem{LiZHZW19}
R.~Li, C.~Zhang, Q.~Hu, P.~Zhu, and Z.~Wang, ``Flexible multi-view
  representation learning for subspace clustering,'' in \emph{Proceedings of
  the Twenty-Eighth International Joint Conference on Artificial Intelligence,
  {IJCAI} 2019, Macao, China, August 10-16, 2019}, 2019, pp. 2916--2922.

\bibitem{LiuWZLZLLDY20}
X.~Liu, L.~Wang, X.~Zhu, M.~Li, E.~Zhu, T.~Liu, L.~Liu, Y.~Dou, and J.~Yin,
  ``Absent multiple kernel learning algorithms,'' \emph{{IEEE} Trans. Pattern
  Anal. Mach. Intell.}, vol.~42, no.~6, pp. 1303--1316, 2020.

\bibitem{liu2018late}
X.~Liu, X.~Zhu, M.~Li, L.~Wang, C.~Tang, J.~Yin, D.~Shen, H.~Wang, and W.~Gao,
  ``Late fusion incomplete multi-view clustering,'' \emph{IEEE Transactions on
  Pattern Analysis and Machine Intelligence}, vol.~41, no.~10, pp. 2410--2423,
  2018.

\bibitem{2020Adaptive}
J.~Wen, K.~Yan, Z.~Zhang, Y.~Xu, and B.~Zhang, ``Adaptive graph completion
  based incomplete multi-view clustering,'' \emph{IEEE Transactions on
  Multimedia}, vol.~23, no.~99, pp. 2493--2504, 2020.

\bibitem{Lin_2021_CVPR}
Y.~Lin, Y.~Gou, Z.~Liu, B.~Li, J.~Lv, and X.~Peng, ``Completer: Incomplete
  multi-view clustering via contrastive prediction,'' in \emph{Proceedings of
  the IEEE/CVF Conference on Computer Vision and Pattern Recognition (CVPR)},
  June 2021, pp. 11\,174--11\,183.

\bibitem{Tran0ZJ17}
L.~Tran, X.~Liu, J.~Zhou, and R.~Jin, ``Missing modalities imputation via
  cascaded residual autoencoder,'' in \emph{2017 {IEEE} Conference on Computer
  Vision and Pattern Recognition, {CVPR} 2017, Honolulu, HI, USA, July 21-26,
  2017}, 2017, pp. 4971--4980.

\bibitem{MarlinZRS11}
B.~M. Marlin, R.~S. Zemel, S.~T. Roweis, and M.~Slaney, ``Recommender systems,
  missing data and statistical model estimation,'' in \emph{{IJCAI} 2011,
  Proceedings of the 22nd International Joint Conference on Artificial
  Intelligence, Barcelona, Catalonia, Spain, July 16-22, 2011}, 2011, pp.
  2686--2691.

\bibitem{WangZLYZ19}
H.~Wang, L.~Zong, B.~Liu, Y.~Yang, and W.~Zhou, ``Spectral perturbation meets
  incomplete multi-view data,'' in \emph{Proceedings of the Twenty-Eighth
  International Joint Conference on Artificial Intelligence, {IJCAI} 2019,
  Macao, China, August 10-16, 2019}, 2019, pp. 3677--3683.

\bibitem{yan2021deep}
X.~Yan, S.~Hu, Y.~Mao, Y.~Ye, and H.~Yu, ``Deep multi-view learning methods: A
  review,'' \emph{Neurocomputing}, 2021.

\bibitem{KumarD11}
A.~Kumar and H.~D. III, ``A co-training approach for multi-view spectral
  clustering,'' in \emph{Proceedings of the 28th International Conference on
  Machine Learning, {ICML} 2011, Bellevue, Washington, USA, June 28 - July 2,
  2011}, 2011, pp. 393--400.

\bibitem{AppiceM16}
A.~Appice and D.~Malerba, ``A co-training strategy for multiple view clustering
  in process mining,'' \emph{{IEEE} Trans. Serv. Comput.}, vol.~9, no.~6, pp.
  832--845, 2016.

\bibitem{SunLXB15}
J.~Sun, J.~Lu, T.~Xu, and J.~Bi, ``Multi-view sparse co-clustering via proximal
  alternating linearized minimization,'' in \emph{Proceedings of the 32nd
  International Conference on Machine Learning, {ICML} 2015, Lille, France,
  6-11 July 2015}, 2015, pp. 757--766.

\bibitem{KumarRD11}
A.~Kumar, P.~Rai, and H.~D. III, ``Co-regularized multi-view spectral
  clustering,'' in \emph{Advances in Neural Information Processing Systems 24:
  25th Annual Conference on Neural Information Processing Systems 2011.
  Proceedings of a Meeting Held 12-14 December 2011, Granada, Spain}, 2011, pp.
  1413--1421.

\bibitem{cca}
H.~Hotelling, ``Relations between two sets of variates,'' in
  \emph{Breakthroughs in Statistics}, 1992, pp. 162--190.

\bibitem{dcca}
G.~Andrew, R.~Arora, J.~A. Bilmes, and K.~Livescu, ``Deep canonical correlation
  analysis,'' in \emph{Proceedings of the 30th International Conference on
  Machine Learning, {ICML} 2013, Atlanta, GA, USA, 16-21 June 2013}, 2013, pp.
  1247--1255.

\bibitem{dccae}
W.~Wang, R.~Arora, K.~Livescu, and J.~A. Bilmes, ``On deep multi-view
  representation learning,'' in \emph{Proceedings of the 32nd International
  Conference on Machine Learning, {ICML} 2015, Lille, France, 6-11 July 2015},
  2015, pp. 1083--1092.

\bibitem{LanckrietCBGJ03}
G.~R.~G. Lanckriet, N.~Cristianini, P.~L. Bartlett, L.~E. Ghaoui, and M.~I.
  Jordan, ``Learning the kernel matrix with semidefinite programming,''
  \emph{J. Mach. Learn. Res.}, vol.~5, pp. 27--72, 2004.

\bibitem{BachLJ04}
F.~R. Bach, G.~R.~G. Lanckriet, and M.~I. Jordan, ``Multiple kernel learning,
  conic duality, and the {SMO} algorithm,'' in \emph{Machine Learning,
  Proceedings of the Twenty-First International Conference {(ICML} 2004),
  Banff, Alberta, Canada, July 4-8, 2004}, 2004.

\bibitem{SonnenburgRSS06}
S.~Sonnenburg, G.~R{\"{a}}tsch, C.~Sch{\"{a}}fer, and B.~Sch{\"{o}}lkopf,
  ``Large scale multiple kernel learning,'' \emph{J. Mach. Learn. Res.},
  vol.~7, pp. 1531--1565, 2006.

\bibitem{SimpleMKL}
A.~Rakotomamonjy, F.~Bach, S.~Canu, and Y.~Grandvalet, ``{SimpleMKL},''
  \emph{{Journal of Machine Learning Research}}, vol.~9, pp. 2491--2521, 2008.

\bibitem{2018Late}
X.~Liu, X.~Zhu, M.~Li, L.~Wang, C.~Tang, J.~Yin, D.~Shen, H.~Wang, and W.~Gao,
  ``Late fusion incomplete multi-view clustering,'' \emph{IEEE Transactions on
  Pattern Analysis and Machine Intelligence}, vol.~PP, pp. 1--1, 2018.

\bibitem{comlet1}
C.~Xu, D.~Tao, and C.~Xu, ``Multi-view learning with incomplete views,''
  \emph{{IEEE} Trans. Image Process.}, vol.~24, no.~12, pp. 5812--5825, 2015.

\bibitem{comlet2}
W.~Shao, X.~Shi, and P.~S. Yu, ``Clustering on multiple incomplete datasets via
  collective kernel learning,'' in \emph{2013 {IEEE} 13th International
  Conference on Data Mining, Dallas, TX, USA, December 7-10, 2013}, 2013, pp.
  1181--1186.

\bibitem{YuanWTNY12}
L.~Yuan, Y.~Wang, P.~M. Thompson, V.~A. Narayan, and J.~Ye, ``Multi-source
  learning for joint analysis of incomplete multi-modality neuroimaging data,''
  in \emph{The 18th {ACM} {SIGKDD} International Conference on Knowledge
  Discovery and Data Mining, {KDD} '12, Beijing, China, August 12-16, 2012},
  2012, pp. 1149--1157.

\bibitem{pvc}
S.~Li, Y.~Jiang, and Z.~Zhou, ``Partial multi-view clustering,'' in
  \emph{Proceedings of the Twenty-Eighth {AAAI} Conference on Artificial
  Intelligence, July 27 -31, 2014, Qu{\'{e}}bec City, Qu{\'{e}}bec, Canada},
  2014, pp. 1968--1974.

\bibitem{img}
H.~Zhao, H.~Liu, and Y.~Fu, ``Incomplete multi-modal visual data grouping,'' in
  \emph{Proceedings of the Twenty-Fifth International Joint Conference on
  Artificial Intelligence, {IJCAI} 2016, New York, NY, USA, 9-15 July 2016},
  2016, pp. 2392--2398.

\bibitem{gans}
Q.~Wang, Z.~Ding, Z.~Tao, Q.~Gao, and Y.~Fu, ``Partial multi-view clustering
  via consistent {GAN},'' in \emph{{IEEE} International Conference on Data
  Mining, {ICDM} 2018, Singapore, November 17-20, 2018}, 2018, pp. 1290--1295.

\bibitem{group1}
H.~Zhao, H.~Liu, and Y.~Fu, ``Incomplete multi-modal visual data grouping,'' in
  \emph{Proceedings of the Twenty-Fifth International Joint Conference on
  Artificial Intelligence, {IJCAI} 2016, New York, NY, USA, 9-15 July 2016},
  2016, pp. 2392--2398.

\bibitem{latent1}
Q.~Yin, S.~Wu, and L.~Wang, ``Unified subspace learning for incomplete and
  unlabeled multi-view data,'' \emph{Pattern Recognit.}, vol.~67, pp. 313--327,
  2017.

\bibitem{latent2}
L.~Zhao, Z.~Chen, Y.~Yang, Z.~J. Wang, and V.~C.~M. Leung, ``Incomplete
  multi-view clustering via deep semantic mapping,'' \emph{Neurocomputing},
  vol. 275, pp. 1053--1062, 2018.

\bibitem{cpm}
C.~Zhang, Z.~Han, Y.~Cui, H.~Fu, J.~T. Zhou, and Q.~Hu, ``Cpm-nets: Cross
  partial multi-view networks,'' in \emph{Advances in Neural Information
  Processing Systems 32: Annual Conference on Neural Information Processing
  Systems 2019, NeurIPS 2019, December 8-14, 2019, Vancouver, BC, Canada},
  2019, pp. 557--567.

\bibitem{Heterogeneous}
X.~Wang, H.~Ji, C.~Shi, B.~Wang, Y.~Ye, P.~Cui, and P.~S. Yu, ``Heterogeneous
  graph attention network,'' in \emph{The World Wide Web Conference, {WWW}
  2019, San Francisco, CA, USA, May 13-17, 2019}, 2019, pp. 2022--2032.

\bibitem{ChangHTQAH15}
S.~Chang, W.~Han, J.~Tang, G.~Qi, C.~C. Aggarwal, and T.~S. Huang,
  ``Heterogeneous network embedding via deep architectures,'' in
  \emph{Proceedings of the 21th {ACM} {SIGKDD} International Conference on
  Knowledge Discovery and Data Mining, Sydney, NSW, Australia, August 10-13,
  2015}, 2015, pp. 119--128.

\bibitem{ZhangXKLMZ18}
Y.~Zhang, Y.~Xiong, X.~Kong, S.~Li, J.~Mi, and Y.~Zhu, ``Deep collective
  classification in heterogeneous information networks,'' in \emph{Proceedings
  of the 2018 World Wide Web Conference on World Wide Web, {WWW} 2018, Lyon,
  France, April 23-27, 2018}, 2018, pp. 399--408.

\bibitem{KipfW17}
T.~N. Kipf and M.~Welling, ``Semi-supervised classification with graph
  convolutional networks,'' in \emph{5th International Conference on Learning
  Representations, {ICLR} 2017, Toulon, France, April 24-26, 2017, Conference
  Track Proceedings}, 2017.

\bibitem{gat}
P.~Velickovic, G.~Cucurull, A.~Casanova, A.~Romero, P.~Li, and Y.~Bengio,
  ``Graph attention networks,'' \emph{CoRR}, vol. abs/1710.10903, 2017.

\bibitem{graphsage}
W.~L. Hamilton, Z.~Ying, and J.~Leskovec, ``Inductive representation learning
  on large graphs,'' in \emph{{NIPS}}, 2017, pp. 1024--1034.

\bibitem{hat}
X.~Wang, H.~Ji, C.~Shi, B.~Wang, Y.~Ye, P.~Cui, and P.~S. Yu, ``Heterogeneous
  graph attention network,'' in \emph{The World Wide Web Conference}, 2019, pp.
  2022--2032.

\bibitem{SchlichtkrullKB18}
M.~S. Schlichtkrull, T.~N. Kipf, P.~Bloem, R.~van~den Berg, I.~Titov, and
  M.~Welling, ``Modeling relational data with graph convolutional networks,''
  in \emph{The Semantic Web - 15th International Conference, {ESWC} 2018,
  Heraklion, Crete, Greece, June 3-7, 2018, Proceedings}, ser. Lecture Notes in
  Computer Science, 2018, pp. 593--607.

\bibitem{zhang2019heterogeneous}
C.~Zhang, D.~Song, C.~Huang, A.~Swami, and N.~V. Chawla, ``Heterogeneous graph
  neural network,'' in \emph{Proceedings of the 25th ACM SIGKDD International
  Conference on Knowledge Discovery \& Data Mining}, 2019, pp. 793--803.

\bibitem{partial}
C.~Zhang, Y.~Cui, Z.~Han, J.~T. Zhou, H.~Fu, and Q.~Hu, ``Deep partial
  multi-view learning,'' \emph{IEEE Transactions on Pattern Analysis and
  Machine Intelligence, doi:10.1109/TPAMI.2020.3037734}, 2020.

\bibitem{dmf}
H.~Zhao, Z.~Ding, and Y.~Fu, ``Multi-view clustering via deep matrix
  factorization,'' in \emph{Proceedings of the Thirty-First {AAAI} Conference
  on Artificial Intelligence, February 4-9, 2017, San Francisco, California,
  {USA}}, 2017, pp. 2921--2927.

\bibitem{mdcr}
C.~Zhang, H.~Fu, Q.~Hu, P.~Zhu, and X.~Cao, ``Flexible multi-view
  dimensionality co-reduction,'' \emph{{IEEE} Trans. Image Process.}, vol.~26,
  no.~2, pp. 648--659, 2017.

\bibitem{itml}
J.~V. Davis, B.~Kulis, P.~Jain, S.~Sra, and I.~S. Dhillon,
  ``Information-theoretic metric learning,'' in \emph{Machine Learning,
  Proceedings of the Twenty-Fourth International Conference {(ICML} 2007),
  Corvallis, Oregon, USA, June 20-24, 2007}, 2007, pp. 209--216.

\bibitem{lmnn}
K.~Q. Weinberger and L.~K. Saul, ``Distance metric learning for large margin
  nearest neighbor classification.'' \emph{Journal of Machine Learning
  Research}, vol.~10, no.~2, 2009.

\bibitem{ueaf}
J.~Wen, Z.~Zhang, Y.~Xu, B.~Zhang, L.~Fei, and H.~Liu, ``Unified embedding
  alignment with missing views inferring for incomplete multi-view
  clustering,'' in \emph{Proceedings of the AAAI Conference on Artificial
  Intelligence}, vol.~33, no.~01, 2019, pp. 5393--5400.

\bibitem{lin2021completer}
Y.~Lin, Y.~Gou, Z.~Liu, B.~Li, J.~Lv, and X.~Peng, ``Completer: Incomplete
  multi-view clustering via contrastive prediction,'' in \emph{Proceedings of
  the IEEE/CVF Conference on Computer Vision and Pattern Recognition}, 2021,
  pp. 11\,174--11\,183.

\bibitem{georghiades2001few}
A.~S. Georghiades, P.~N. Belhumeur, and D.~J. Kriegman, ``From few to many:
  Illumination cone models for face recognition under variable lighting and
  pose,'' \emph{IEEE Transactions on Pattern Analysis and Machine
  Intelligence}, vol.~23, no.~6, pp. 643--660, 2001.

\bibitem{wah2011caltech}
C.~Wah, S.~Branson, P.~Welinder, P.~Perona, and S.~Belongie, ``The caltech-ucsd
  birds-200-2011 dataset,'' 2011.

\bibitem{le2014distributed}
Q.~Le and T.~Mikolov, ``Distributed representations of sentences and
  documents,'' in \emph{International Conference on Machine Learning}.\hskip
  1em plus 0.5em minus 0.4em\relax PMLR, 2014, pp. 1188--1196.

\bibitem{22}
C.~H. Lampert, H.~Nickisch, and S.~Harmeling, ``Attribute-based classification
  for zero-shot visual object categorization,'' \emph{IEEE Transactions on
  Pattern Analysis and Machine Intelligence}, vol.~36, no.~3, pp. 453--465,
  2013.

\bibitem{decaf}
A.~Krizhevsky, I.~Sutskever, and G.~E. Hinton, ``Imagenet classification with
  deep convolutional neural networks,'' in \emph{Advances in Neural Information
  Processing Systems 25: 26th Annual Conference on Neural Information
  Processing Systems 2012. Proceedings of a meeting held December 3-6, 2012,
  Lake Tahoe, Nevada, United States}, 2012, pp. 1106--1114.

\bibitem{vgg}
K.~Simonyan and A.~Zisserman, ``Very deep convolutional networks for
  large-scale image recognition,'' in \emph{3rd International Conference on
  Learning Representations, {ICLR} 2015, San Diego, CA, USA, May 7-9, 2015,
  Conference Track Proceedings}, 2015.

\bibitem{2015Large}
Y.~Li, F.~Nie, H.~Huang, and J.~Huang, ``Large-scale multi-view spectral
  clustering via bipartite graph,'' in \emph{Twenty-Ninth AAAI Conference on
  Artificial Intelligence}, 2015.

\bibitem{yang2010bag}
Y.~Yang and S.~Newsam, ``Bag-of-visual-words and spatial extensions for
  land-use classification,'' in \emph{Proceedings of the 18th SIGSPATIAL
  International Conference on Advances in Geographic Information Systems},
  2010, pp. 270--279.

\bibitem{2005A}
F.~Li and P.~Perona, ``A bayesian hierarchical model for learning natural scene
  categories,'' in \emph{Proc. IEEE Computer Society Conference on Computer
  Vision and Pattern Recgnition, May 2005}, 2005.

\bibitem{tsne}
L.~Van~der Maaten and G.~Hinton, ``Visualizing data using t-sne.''
  \emph{Journal of Machine Learning Research}, vol.~9, no.~11, 2008.

\bibitem{CRA}
L.~Tran, X.~Liu, J.~Zhou, and R.~Jin, ``Missing modalities imputation via
  cascaded residual autoencoder,'' in \emph{Proceedings of the IEEE Conference
  on Computer Vision and Pattern Recognition}, 2017, pp. 1405--1414.

\bibitem{9115215}
H.~Huang, J.~Zhang, J.~Zhang, J.~Xu, and Q.~Wu, ``Low-rank pairwise alignment
  bilinear network for few-shot fine-grained image classification,'' \emph{IEEE
  Transactions on Multimedia}, vol.~23, pp. 1666--1680, 2021.

\bibitem{9326401}
J.~Li, W.~Shi, Q.~Ye, N.~Zhang, W.~Zhuang, and X.~Shen, ``Multiservice function
  chain embedding with delay guarantee: A game-theoretical approach,''
  \emph{IEEE Internet of Things Journal}, vol.~8, no.~14, pp. 11\,219--11\,232,
  2021.

\end{thebibliography}
\vspace{-10mm}
\begin{IEEEbiography}[{\includegraphics[width=1in,height=1.25in,clip,keepaspectratio]{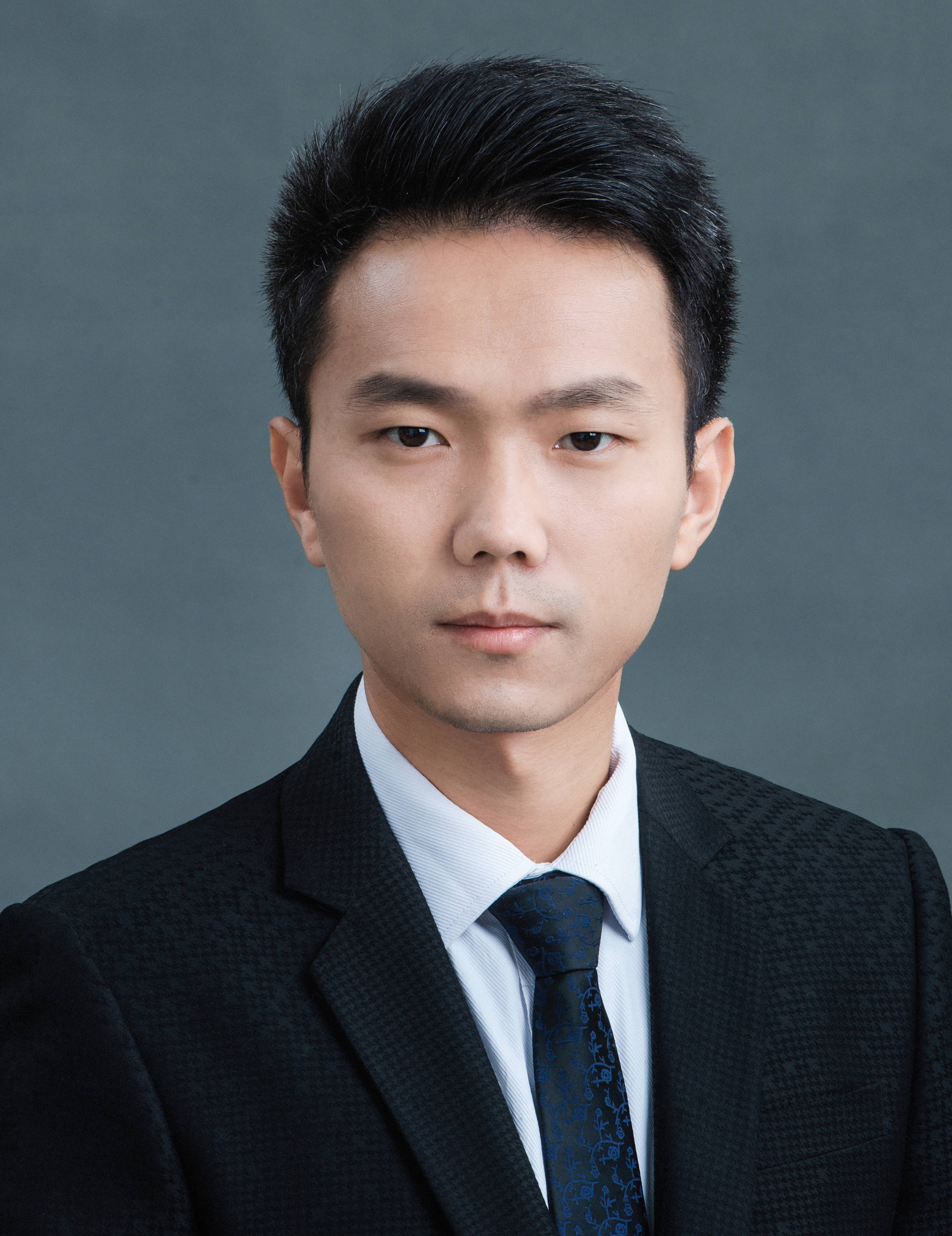}}]{Pengfei Zhu} received the Ph.D. degree from The Hong Kong Polytechnic University, Hong Kong, China, in 2015. Now he is an associate professor with the College of Intelligence and Computing, Tianjin University. His research interests are focused on machine learning and computer vision.
\end{IEEEbiography}
\vspace{-10mm}

\begin{IEEEbiography}[{\includegraphics[width=1in,height=1.25in,clip,keepaspectratio]{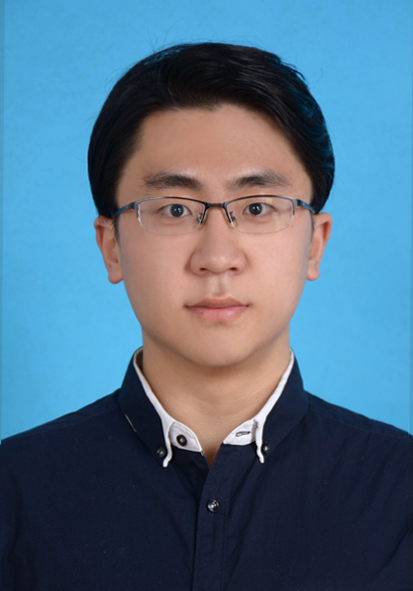}}]{Xinjie Yao} received the B.S. degree in information security from North China Electric Power University, Baoding, China, in 2018, and the M.S. degree in computer technology from China University of Petroleum (East China), Qingdao, China, in 2021. He is currently pursuing the Ph.D. degree in computer science and technology with Tianjin University, Tianjin, China. His research interests are focused on machine learning and computer vision.
\end{IEEEbiography}
\vspace{-10mm}

\begin{IEEEbiography}[{\includegraphics[width=1in,height=1.25in,clip,keepaspectratio]{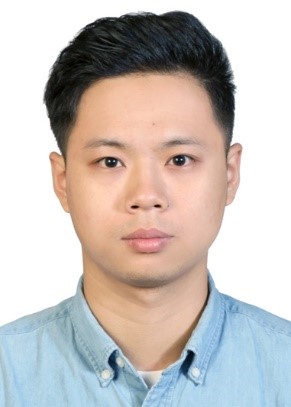}}]{Yu Wang} received the B.S. degree in communication engineering, the M.S. degree in software engineering, and Ph.D. degree in computer applications and techniques from Tianjin University in 2013 and 2016, and 2020, respectively. He is currently an assistant professor of Tianjin University. His research interests focus on hierarchical learning and large-scale classification in industrial scenarios and computer vision applications, data mining and machine learning.

\end{IEEEbiography}

\vspace{-10mm}

\begin{IEEEbiography}[{\includegraphics[width=1in,height=1.25in,clip,keepaspectratio]{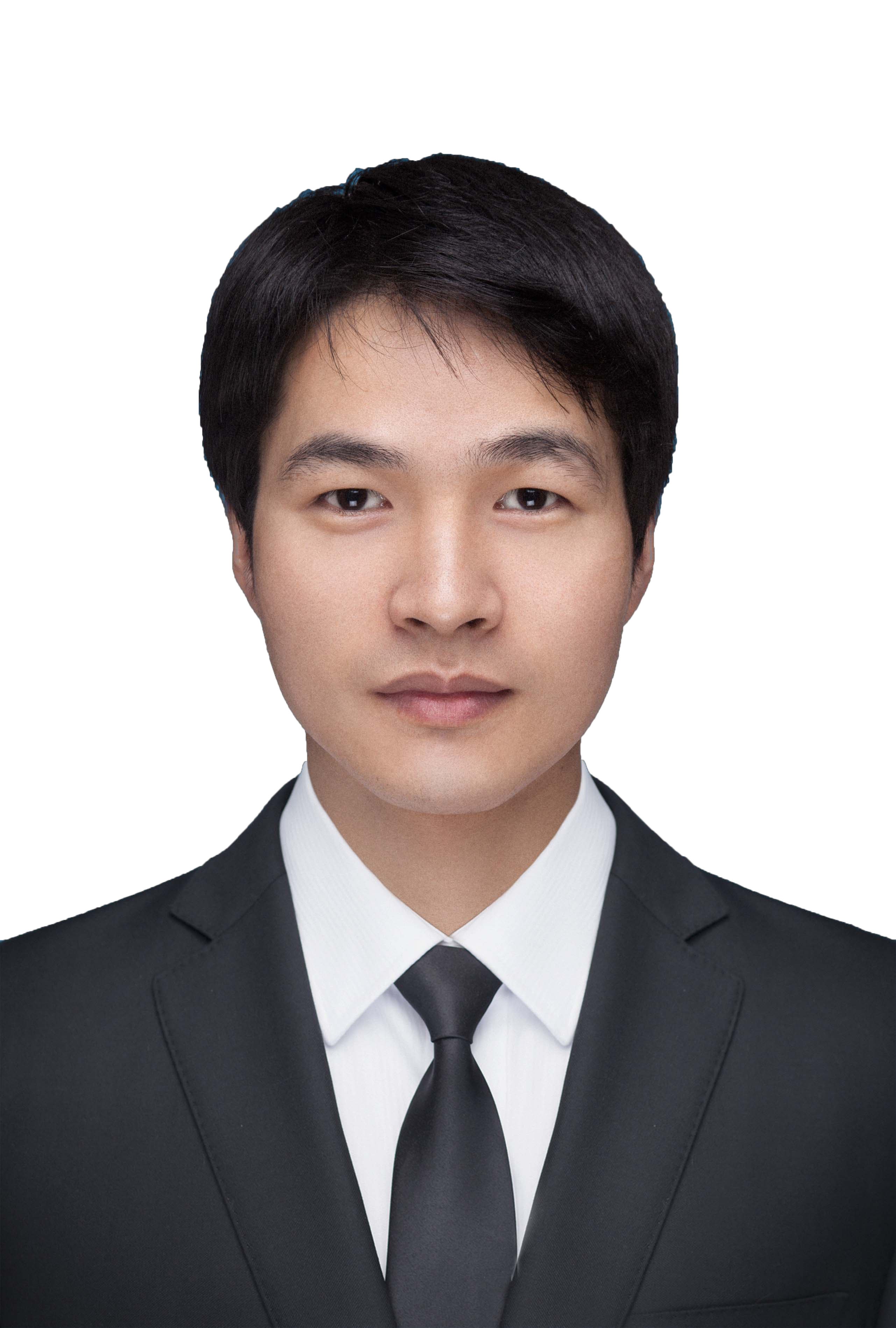}}]{Meng Cao} received the B.S. degree in computer science and technology from Northeastern University, China. He is currently a master student in the College of Intelligence and Computing, Tianjin University. His research interests include pattern recognition and multi-view learning.
\end{IEEEbiography}

\vspace{-10mm}

\begin{IEEEbiography}[{\includegraphics[width=0.9in,height=1.25in,clip,keepaspectratio]{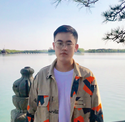}}]{Binyuan Hui} received the M.S. degree in computer science and technology from the College of Intelligence and Computing, Tianjin University, China, in 2020. His research interests include data mining, pattern recognition, computer vision, and natural language processing.
\end{IEEEbiography}

\vspace{-10mm}

\begin{IEEEbiography}[{\includegraphics[width=1in,height=1.25in,clip,keepaspectratio]{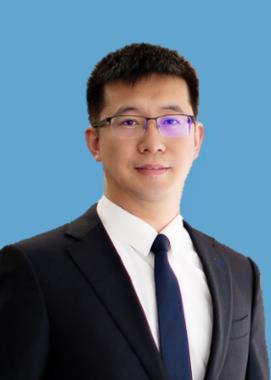}}]{Shuai Zhao} received the B.S. degree in Vehicle Engineering, the the M.S. degree in Vehicle Engineering from Jilin University in 2011 and 2014 respectively. Now He is pursuing the Ph.D. degree in computer applications and techniques in Tianjin University. 
In automotive industry, he works as the director of the ICV data department of China Automotive Data Co., Ltd. (CATARC-ADC), the Coordination Expert of international standards and regulations of China Automotive Standards Committee (CASC), Member of C-ASAM Steering Committee. He is mainly engaged in the related work of ICV scenario database and simulation testing, and participated in the formulation of standards and policies for intelligent vehicle simulation at home and abroad.
\end{IEEEbiography}

\vspace{-10mm}

\begin{IEEEbiography}[{\includegraphics[width=1in,height=1.25in,clip,keepaspectratio]{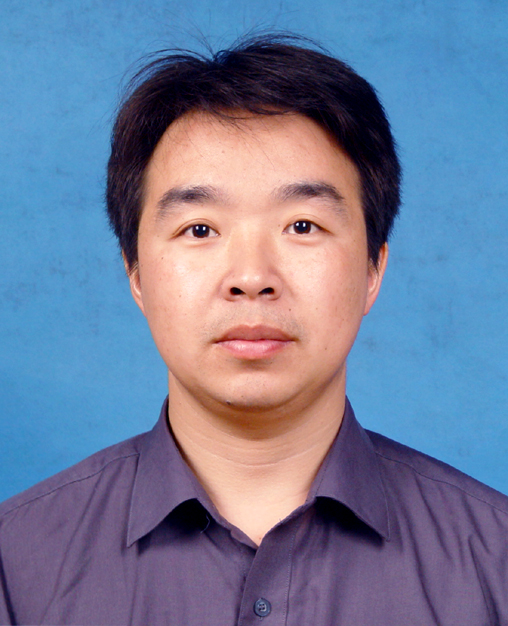}}]{Qinghua Hu} received the B.S., M.S., and Ph.D. degrees from the Harbin Institute of Technology, Harbin, China, in 1999, 2002, and 2008, respectively. He has published over 200 peer-reviewed papers. His current research is focused on uncertainty modeling in big data, machine learning with multi-modality data, intelligent unmanned systems. He is an Associate Editor of the IEEE TRANSACTIONS ON FUZZY SYSTEMS, Acta Automatica Sinica, and Energies.
\end{IEEEbiography}

\end{document}